\documentclass{article}

\usepackage{microtype}
\usepackage{graphicx}
\usepackage{subfigure}
\usepackage{booktabs} 

\usepackage{hyperref}

\usepackage[accepted]{icml2025}

\usepackage{amsmath}
\usepackage{amssymb}
\usepackage{mathtools}
\usepackage{amsthm}

\usepackage[capitalize,noabbrev]{cleveref}

\theoremstyle{plain}

\theoremstyle{definition}

\theoremstyle{remark}

\usepackage[textsize=tiny]{todonotes}

\usepackage[utf8]{inputenc} 
\usepackage[T1]{fontenc}    
\usepackage{url}            
\usepackage{amsfonts}       
\usepackage{nicefrac}       
\usepackage{xcolor}         
\usepackage{multirow} 
\usepackage{makecell}
\usepackage{ulem}
\usepackage{color, colortbl}
\definecolor{Gray}{gray}{0.9}

\definecolor{00red}{RGB}{236,35,35}

\definecolor{00blue}{RGB}{50,149,237}

\usepackage{bbding}
\usepackage{multirow}
\usepackage{subcaption}
\usepackage{indentfirst}
\usepackage{graphicx}
\usepackage{fontawesome}
\usepackage{makecell}
\newcommand{\ie}{\textit{i.e.}}

\newcommand{\eg}{\textit{e.g.}}

\definecolor{mygray}{gray}{0.95}

\newcommand\blfootnote[1]{%
  \begingroup
  \renewcommand\thefootnote{}\footnote{#1}%
  \addtocounter{footnote}{-1}%
  \endgroup
}

\icmltitlerunning{Mulberry: Empowering MLLM with o1-like Reasoning and Reflection via Collective Monte Carlo Tree Search}

\begin{document}

\twocolumn[
\icmltitle{\includegraphics[height=20pt,width=20pt]{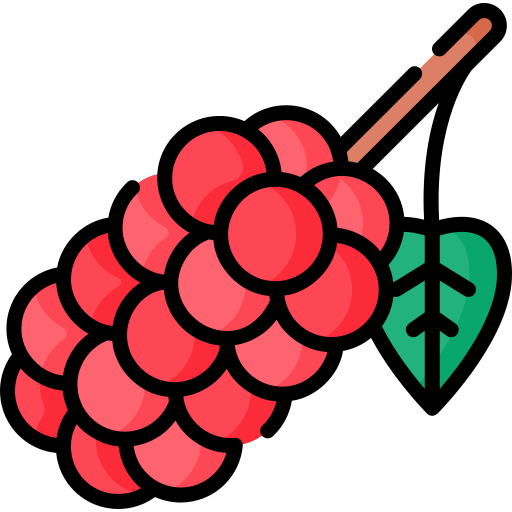} Mulberry: Empowering MLLM with o1-like Reasoning and Reflection via Collective Monte Carlo Tree Search}



\icmlsetsymbol{equal}{*}
\icmlsetsymbol{corresponding}{\textsuperscript{\Envelope}}

\begin{icmlauthorlist}
\icmlauthor{Huanjin Yao$^{2, 3, *}$}{}
\icmlauthor{Jiaxing Huang$^{1, *, \text{\Envelope}}$}{}
\icmlauthor{Wenhao Wu$^3$}{}
\icmlauthor{Jingyi Zhang$^1$}{}
\icmlauthor{Yibo Wang$^2$}{}
\icmlauthor{Shunyu Liu$^1$}{}
\icmlauthor{Yingjie Wang$^1$}{}
\icmlauthor{Yuxin Song$^3$}{}
\icmlauthor{Haocheng Feng$^3$}{}
\icmlauthor{Li Shen$^4$}{}
\icmlauthor{Dacheng Tao$^1$}{}
\end{icmlauthorlist}

\icmlcorrespondingauthor{Firstname1 Lastname1}{first1.last1@xxx.edu}
\icmlcorrespondingauthor{Firstname2 Lastname2}{first2.last2@www.uk}

\icmlkeywords{Machine Learning, ICML}

\vskip 0.3in
]

\blfootnote{
$^*$ Equal Contribution.
Correspondence to: Jiaxing Huang <jiaxing.huang@ntu.edu.sg>.
$^1$ Nanyang Technological
University; $^2$ Tsinghua University; $^3$ Baidu Inc.; $^4$ Sun Yat-sen
University.
} 

\begin{abstract}
In this work, we aim to develop an MLLM that understands and solves questions by learning to create each intermediate step of the reasoning involved till the final answer.
To this end, we propose Collective Monte Carlo Tree Search (CoMCTS), a new learning-to-reason method for MLLMs, which introduces the concept of collective learning into ``tree search'' for effective and efficient reasoning-path searching and learning. 
The core idea of CoMCTS is to leverage 
collective knowledge from multiple models
to collaboratively conjecture, search and identify effective reasoning paths toward correct answers via four iterative operations including Expansion, Simulation and Error Positioning, Backpropagation, and Selection.
Using CoMCTS, we construct Mulberry-260k, a multimodal dataset with a tree of rich, explicit and well-defined reasoning nodes for each question.
With Mulberry-260k, we perform collective SFT to train our model, Mulberry, a series of MLLMs with o1-like step-by-step Reasoning and Reflection capabilities.
Extensive experiments demonstrate the superiority of our proposed methods on various benchmarks.
Code will be available at \url{https://github.com/HJYao00/Mulberry}

\end{abstract}

\section{Introduction}
\label{sec:intro}

\begin{figure}[t]
\centering
\includegraphics[width=1\linewidth]{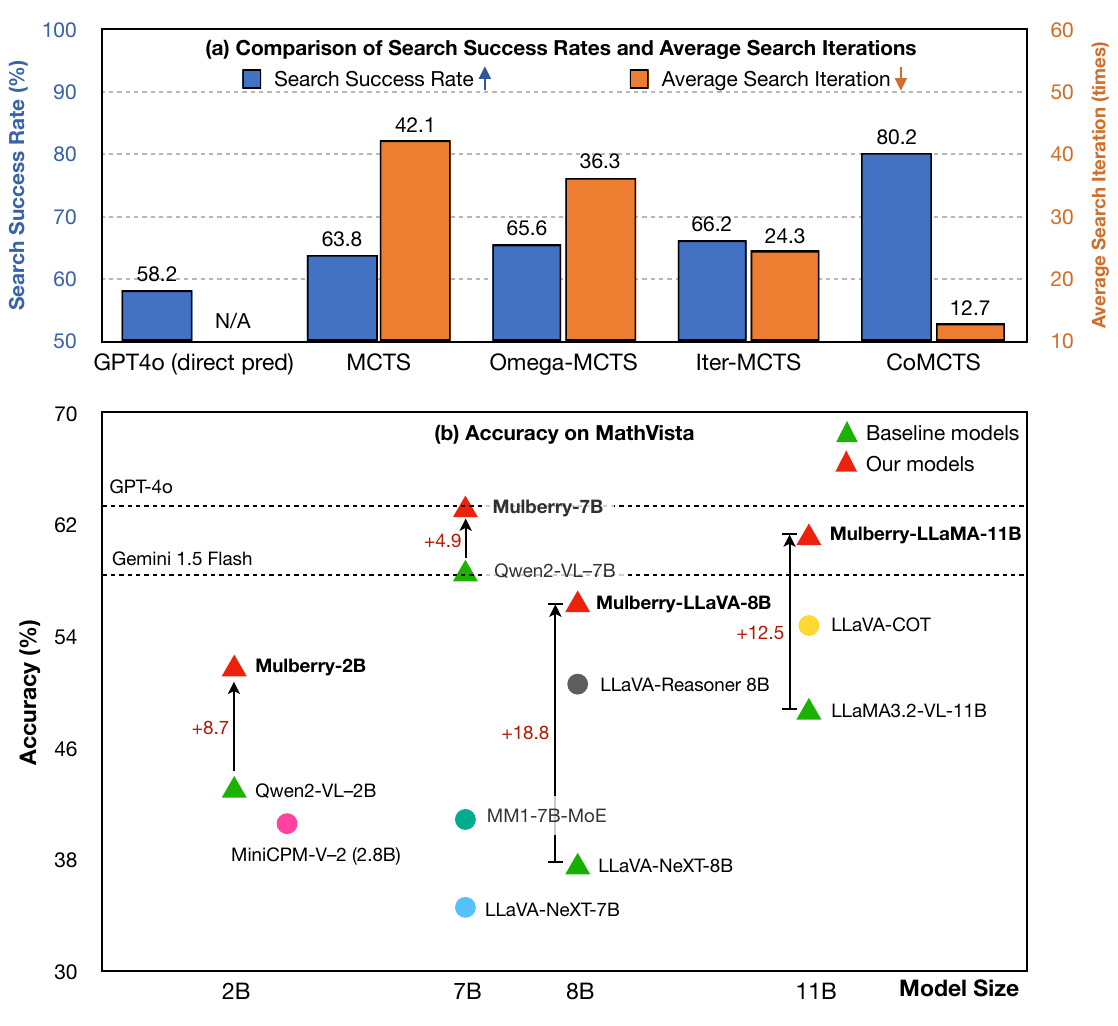}
\caption{
\textbf{(a)} Our CoMCTS shows great superiority in search effectiveness and efficiency against other tree search methods.
\textbf{(b)} Our Mulberry, trained on CoMCTS-searched data, outperforms most open-sourced MLLMs and achieves competitive results against closed-source ones, showing outstanding abilities in step-by-step reasoning and reflection.
}
\label{fig:1}

\end{figure}

\textit{``What I cannot create, I do not understand.''} \\
\rightline{\textit{---Richard Feynman}}

Multimodal large language models (MLLMs) embody the essence of this dictum, 
which understand the world by learning to create expected responses to multimodal inputs such as images and text.
While MLLMs have recently shown significant progress in straightforward tasks~\cite{llava,qwen2vl}, they often experience obviously increased failures on complex tasks requiring in-depth reasoning~\cite{llava-reasoner}.
Feynman’s dictum might be the perfect metaphor of such failures of MLLMs, as we should only be able to work something out if we 
can create and 
have a firm understanding of each step of the reasoning involved. 
However, current MLLMs predominantly operate in a simple ``direct prediction'' mode~\cite{llava-cot}, \ie,  generating brief, final answers to questions with little explicit and well-defined intermediate reasoning steps.

In this work, we aim to develop an MLLM that understands and solves questions by learning to create each intermediate step of the reasoning involved till the final answer. 
Recent advances in NLP, such as OpenAI o1~\cite{openai2024o1}, have shown great potential in enabling LLM to learn to reason and tackle complex language tasks~\cite{xie2024monte}. 
The core design of these advances lies in AlphaGo-like ``tree search'': they employ tree search methods, like MCTS~\cite{coulom2006efficient}, to bootstrap an LLM itself to build a tree of intermediate thoughts, explore effective reasoning paths, and leverage these paths to teach the model to reason step-by-step.

An intuitive idea is to directly apply these ``tree search'' methods to search effective reasoning paths for MLLMs, which, however, does not work well. 
As illustrated in Figure~\ref{fig:1}, we believe this is largely attributed to several observed search challenges for MLLMs.
(1)	\textit{Search Effectiveness:} 
Traditional MCTS methods generally work by self-bootstrapping while current MLLMs are typically trained with little explicit and well-defined intermediate reasoning steps, making these search methods often trapped in homogeneous low-quality nodes within the reasoning space of a single MLLM, ultimately leading to low search success rates.
(2)	\textit{Search Efficiency:} Traditional MCTS methods typically expand and explore only one subsequent reasoning node per search iteration, which advance a single step each time and demand massive iterations, making them inefficient for computation-intensive MLLMs.

To tackle these challenges, we propose Collective Monte Carlo Tree Search (CoMCTS), a new learning-to-reason method for MLLMs, which introduces the concept of collective learning into ``tree search'' for effective and efficient reasoning-path searching and learning.
The core idea of CoMCTS is to leverage collective knowledge to collaboratively conjecture, search and identify effective reasoning paths toward correct answers.
Specifically, CoMCTS searches effective reasoning paths iteratively, and in each iteration, it leverages collective knowledge from multiple MLLMs to jointly (a) expand diverse and complementary candidate subsequent reasoning nodes till the end from a given start node, (b) simulate reasoning outcomes, position error candidate nodes and prune them along with their child nodes, (c) backpropagate to update the score and visit count of each reasoning node in a bottom-up manner, and (d) select the leaf reasoning node with the highest Upper Confidence Bound value as next start node.

In this way, our CoMCTS achieves effective and efficient reasoning search. (1) The joint expansion mechanism enables CoMCTS to concatenate reasoning trajectories from multiple MLLMs via iterative search, ultimately constructing an unified
reasoning tree comprising diverse and complementary reasoning nodes. Thus, it allows reasoning-path search not only within the reasoning space of a given MLLM itself but also among those of others, benefiting from the synergy of multiple MLLMs while avoiding being trapped in homogeneous low-quality nodes within the reasoning space of a single MLLM itself.
(2) The joint simulation and error positioning mechanism enables CoMCTS to, in each search iteration, skip multiple intermediate steps and select the last correct step as the next start node, largely reducing search time while 
maintaining search effectiveness.   
Here, collective knowledge is also crucial as it is often challenging for a model to recognize and position errors made by itself while relatively easy by using other models.

Furthermore, we extend our CoMCTS for reflective reasoning-path search.
Based on the unified reasoning tree constructed by CoMCTS, which provides 
both positive and negative reasoning nodes
, we identify and integrate negative sibling nodes into effective reasoning paths to build the reflective reasoning path that includes a transition from a negative reasoning node to a positive one.
By learning from 
reflective reasoning paths, MLLMs can perform appropriate step-wise reflection, dynamically calibrating their reasoning trajectory from an erroneous node toward a correct one during long-chain reasoning.
Here, collective knowledge facilitates reflective reasoning-path search by providing a rich set of diverse positive and negative reasoning nodes.

Using our CoMCTS, we search effective and reflective reasoning paths for a set of multimodal inputs, and construct Mulberry-260k, a Multimodal learning-to-Reason-and-Reflect dataset with a tree of rich, explicit and well-defined reasoning nodes for each question. 
With Mulberry-260k, we perform collective supervised fine-tuning to train our model, Mulberry, a series of Multimodal LLMs with o1-like step-by-step Reasoning and Reflection capabilities.

The main contributions of this work are fourfold. \textit{\textbf{First}}, we introduce the concept of collective learning into MCTS, and propose CoMCTS which leverages collective knowledge to collaboratively conjecture, search and identify effective and reflective reasoning paths for MLLMs, showing great superiority in search effectiveness and efficiency.
To the best of our knowledge, this is the first work that explores collective learning with MCTS for MLLMs.
\textit{\textbf{Second}}, we construct Mulberry-260k that provides a valuable resource for advancing research in step-by-step reasoning and reflection in MLLMs.
\textit{\textbf{Third}}, we develop Mulberry, a series of MLLMs with outstanding capabilities in step-by-step reasoning and reflection.
\textit{\textbf{Fourth}}, extensive experiments demonstrate the superiority of our proposed methods on various benchmarks.

\section{Related Works}
\subsection{Multimodal Large Language Model}
MLLMs~\cite{llava,qwen2vl,deepseek-vl, denseconnector} have made notable advancements in general vision-language understanding, enabling them to interpret visual semantics across various domains.
Recent studies~\cite{mmmu_pro, llava-reasoner} explore MLLM reasoning and reveal that directly employing CoT prompt to derive the final answer may result in limited gains or even degradation.
In addition, some studies~\cite{CCot, luan2024textcot} introduce plan-based CoT prompting to guide models to generate intermediate information for predicting final answers.
Recent advances~\cite{llava-cot} attempt structured reasoning with a planed flow of certain pre-defined stages, enhancing the CoT capabilities~\cite{vision_survey} of MLLMs.
Differently, this paper, for the first time, introduces the concept of ``tree search'' into MLLM reasoning and proposes a novel CoMCTS technique to search effective and reflective reasoning paths to train our Mulberry, a series of MLLMs with outstanding capabilities in step-by-step reasoning and reflection.

\subsection{Large Language Model Reasoning}
LLM reasoning methods can be broadly categorized into three types, including prompt-based, plan-based and learning-based reasoning.
Prompt-based methods, like Chain-of-Thought (CoT)~\cite{wei2022chain}, mimic human reasoning by providing a few hand-crafted, step-by-step solutions as references.
Plan-based methods, such as Tree/Graph-of-thought~\cite{yao2024tree,besta2024graph}, predict multiple reasoning paths in a tree or graph manner and take consistent units of thought for thoughtful decision-making.
Learning-based reasoning methods, represented by GPTo1, 
Star~\cite{zelikman2022star}, 
Iter-MCTS~\cite{xie2024monte}
and ReST-MCTS~\cite{zhang2024rest}, first employ tree search approaches, like MCTS, to bootstrap an LLM itself to build a tree of intermediate thoughts, explore effective reasoning paths, and leverage these paths to train the model to reason step-by-step.

\subsection{Monte-Carlo Tree Search}
Monte-Carlo Tree Search (MCTS) is a powerful search paradigm for complex decision making problems and has been extensively explored across diverse fields, including games~\cite{silver2017mastering,ye2021mastering}, robotics~\cite{best2019dec,dam2022monte}, theorem proving~\cite{lample2022hypertree},
matrices multiplication~\cite{fawzi2022discovering}, etc.
For instance, AlphaGo~\cite{silver2017mastering} introduces deep learning into MCTS, achieving superhuman performance in board and video games~\cite{silver2017mastering,ye2021mastering}. Besides, \cite{pitanov2023monte,yang2023integrated} explore MCTS for path finding and train timetabling problems, while \cite{vagadia2024phyplan} integrates MCTS into physics-informed planning networks for robot control. 
In this work, we propose CoMCTS that enables effective and reflective reasoning-path searching and learning on MLLMs.

\subsection{Collective Learning}
Collective learning, also known as Co-training, aims to harness collective intelligence of multiple individuals to improve learning outcomes.
This concept originates in early pioneering studies~\cite{blum1998combining, sun2011robust, yu2011bayesian}, which utilize collective knowledge to address data insufficiency issues in classification learning.
Recent advances introduce collective learning into deep neural networks for efficient and effective deep learning.
For example, \cite{qiao2018deep,saito2018maximum} employ collective knowledge from multiple classifiers to predict more accurate pseudo-labels for semi-supervised classification; \cite{cui2022genco} utilizes collective knowledge from multiple discriminators to enhance image discrimination and generation; and \cite{foerster2016learning} leverages the synergy of multiple models for reinforcement learning.

\section{Methodology}
\label{sec:method}
We first present our proposed CoMCTS that introduces the concept of collective learning into ``tree search'' for effective and efficient reasoning-path searching and learning.
We then illustrate the extension of CoMCTS for reflective reasoning-path search, and describe data construction (\ie, Mulberry-260k) and model training (\ie, Mulberry) using CoMCTS.
More details to be elaborated in the ensuing subsections.

\begin{figure*}[t]
\centering
\includegraphics[width=1\linewidth]{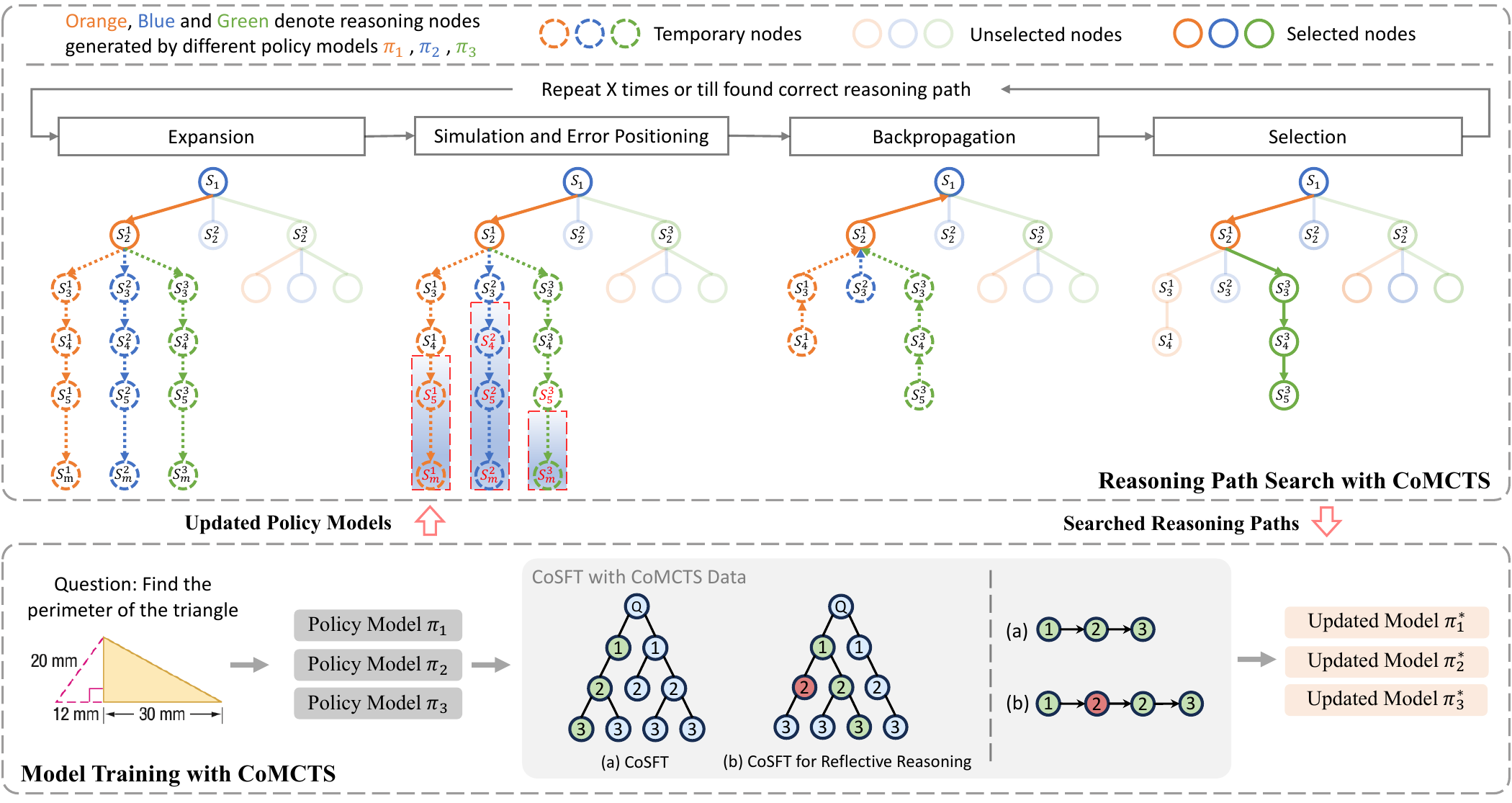}
\caption{
\textbf{Overview.}  
Our CoMCTS trains Mulberry with two alternating phases.
\textit{In top part,} CoMCTS searches reasoning paths iteratively, and in each iteration, it utilizes collective knowledge from multiple MLLMs to jointly (a) expand diverse and complementary candidate subsequent reasoning nodes till the end from a given start node, (b) simulate reasoning outcomes, position error candidate nodes and prune them along with their child nodes, (c) backpropagate to update the score and visit count of each reasoning node in a bottom-up manner, and (d) select the leaf reasoning node with the highest UCB value as next start node. 
\textit{In bottom part,} we train the model to learn from the reasoning trees constructed by CoMCTS.
}
\vspace{-1.0em}
\label{fig:schematic}

\end{figure*}

\subsection{CoMCTS for effective reasoning}
\label{sec: Collective MCTS}

The core idea of CoMCTS is to leverage collective knowledge to collaboratively conjecture, search and identify effective reasoning nodes in an iterative manner, aiming to find effective reasoning paths leading to correct answers.

We denote a policy model as $\pi$, which is initialized by a pre-trained MLLM. 
We leverage collective knowledge from a group of MLLMs $\{\pi_{1}, \pi_{2},...,\pi_{K}\}$ to jointly search and learn effective reasoning paths.
Given a multimodal input question $Q$ (\eg, a text task instruction with an image, $Q = \{\text{text}, \text{image}\} $), each model $\pi$ can generate a sequence of intermediate reasoning states toward the final answer $(s_{1}, s_{2}, s_{3}, ..., s_{M}) \sim \pi_{\theta}(\cdot|Q)$ via autoregressive next token prediction.
We define the intermediate reasoning state at step $m$ as $s_{m}$ and the state generated by model $\pi_{k}$ at step $m$ as $s_{m}^{k}$.
Each reasoning step consists of one or a few sentences containing multiple word tokens.

CoMCTS algorithm begins at the root node, \ie, either the start of a response or an incomplete response, and performs reasoning-path search via a certain number of iterations, where each iteration comprises four key operations: (a) Expansion, (b) Simulation and Error Positioning, (c) Backpropagation, and (d) Selection, as elaborated below.

\textbf{(a) Expansion.} The goal of this operation in CoMCTS is to expand the current leaf reasoning node (if it is not a terminal node) to integrate new subsequent candidate reasoning nodes.
Given the current leaf node $s_{m}^{k}$ (\ie, the node selected by Operation (d) Selection or the root node), CoMCTS utilizes collective knowledge from a group of MLLMs, $\{\pi_{1}, \pi_{2},...,\pi_{K}\}$, to jointly expand a set of diverse and complementary candidate reasoning paths $S_{\text{candidate}} = \cup_{j=1}^{K} S_{\text{candidate}}^{j}$ in parallel till the terminal node: 
\begin{equation}
     S_{\text{candidate}}^{j} \sim \pi_{j}(\cdot|Q, \text{Parent}(s_{m}^{k}), s_{m}^{k}),
\end{equation}
where $\text{Parent}(s_{m}^{k})$ returns all parent nodes of $s_{m}^{k}$ and $(\text{Parent}(s_{m}^{k}), s_{m}^{k})$ denotes the 
current 
reasoning path from the root node to $s_{m}^{k}$. 
${S_{\text{candidate}}^{j}} = \{s_{i}^{j}\}$ stands for a potential reasoning path generated by model $\pi_{j}$ starting from $s_{m}^{k}$.

\textbf{(b) Simulation and Error Positioning.} In this operation, CoMCTS utilizes collective knowledge from $\{\pi_{1}, \pi_{2},... ,\pi_{K}\}$ to jointly estimate the potential value of child nodes $s_{i}^{j} \in S_{\text{candidate}}$ (added in Operation (a)), and considers low-score nodes as erroneous reasoning nodes, and positions and filters out them along with their child nodes:
\begin{equation}
    R(s_{i}^{j}) = \frac{1}{K} \sum_{l=1}^{K} \pi_{l}(\cdot|\text{prompt}_{\text{eval}}, Q, \text{Parent}(s_{i}^{j}), s_{i}^{j})
\end{equation}
\begin{equation}
    S_{\text{candidate}}^{*} = \{s_{i}^{j} \in S_{\text{candidate}} |  R(s_{i}^{j}) >= t\}
\end{equation}
where $R(s_{i}^{j})$ denotes a reasoning node evaluation function that uses the prompt, $\text{prompt}_{\text{eval}}$, to request a group of MLLMs, $\{\pi_{1}, \pi_{2},... ,\pi_{K}\}$, to jointly evaluate the candidate reasoning node $s_{i}^{j}$. 
$t$ is a threshold and Discontinued reasoning nodes in $S_{\text{candidate}}^{*}$ are automatically removed following the error node removal in Eq.(3).

\textbf{(c) Backpropagation.} 
Given the new reasoning tree expanded and simulated using collective knowledge in Operations (a)-(b), CoMCTS performs a bottom-up update from the leaf nodes back to the root node. Each node $s$ along the newly expanded path in the reasoning tree updates its statistics, including visit count $N$ and node value $V$:
\begin{equation}
    V(s) \xleftarrow{} \frac{N(s) \cdot V(s) + \sum_{s_{l} \in \text{Child}(s)} R(s_{l})}{N(s) + \text{CountChild}(S_{\text{candidate}}^{*}, s)},
\end{equation}
\begin{equation}
    N(s) \xleftarrow{} N(s) + \text{CountChild}(S_{\text{candidate}}^{*}, s),
\end{equation}
where $\text{Child}(s)$ returns all the child nodes of $s$, and $\text{CountChild}(S_{\text{candidate}}^{*}, s)$ is a child node counting function that calculates the number of child nodes of $s$ in $S_{\text{candidate}}^{*}$.

\textbf{(d) Selection.}  Following Operations (a), (b) and (c), CoMCTS traverses the updated reasoning tree to select the next starting node. This selection is guided by the Upper Confidence Bound (UCB) value, which balances search exploration and exploitation.
The UCB value of a node $s$ is computed using the node reward value $V(s)$ and the visit cound $N(s)$. Among the candidate nodes $s \in S_{\text{candidate}}^{*}$, the one with the highest UCB value is chosen as the starting node $s_{m}^{k^{*}}$ for next search iteration:

\begin{equation}
    s_{m}^{k^{*}} = \mathop{\arg\max}_{s \in S_{\text{candidate}}^{*}} V(s) + c \cdot \sqrt{\frac{\log N(\hat{s})}{1 + N(s)}}
\label{eq:ucb}
\end{equation}
where $c$ stands for a constant which controls the level of exploration. $\hat{s}$ denotes the parent node of $s$.

\textbf{CoMCTS.} 
These four operations, \ie, (a) Expansion, (b) Simulation and Error Positioning, (c) Backpropagation and (d) Selection, are repeated for a pre-defined number of iterations or until correct reasoning paths are found. This iterative process allows CoMCTS to construct a question-dependent reasoning tree $S$ with the correct reasoning path $Y$, and ultimately form a multimodal learning-to-reason data triplet $\{Q, Y, S\}$. 
By applying our CoMCTS to a set of multimodal questions, we can construct a collection of multimodal learning-to-reason data triplets, which provide a tree of rich, explicit and well-defined reasoning nodes toward the final answer for each question and enable MLLMs to learn to reason step-by-step.

\subsection{CoMCTS for reflective reasoning} 

In this subsection, we extend CoMCTS for reflective reasoning-path search. 
Based on the unified reasoning tree constructed by CoMCTS, \ie, $\{Q, Y, S\}$, which provides both positive and negative reasoning nodes, we identify and integrate negative sibling nodes into effective reasoning paths to build the reflective reasoning path that includes a transition from a negative reasoning node to a positive one.

\textbf{Identifying negative sibling node.} 
Given the effective reasoning path $Y$, we identify the negative sibling reasoning node for $s \in Y$ using UCB:
\begin{equation}
    s_{\text{neg}} = \mathop{\arg\min}_{s_{l} \in \text{Sibling}(s)}  \text{UCB}(s_{l}) - \text{UCB}(s), \ \ \forall s \in Y,
\label{eq:neg_sibling}
\end{equation}
where $\text{Sibling}(s)$ returns all the sibling nodes of $s$, \ie, the nodes on the same hierarchical level under the same parent node of $s$. $\text{UCB}(s) = V(s) + c \cdot \sqrt{\frac{\log N(\hat{s})}{1 + N(s)}}$ as in Eq.~\ref{eq:ucb}.

\textbf{Constructing reflective reasoning path.} 
Based on Eq.~\ref{eq:neg_sibling}, we randomly sample a reasoning node $s \in Y$ with its negative sibling node $s_{\text{neg}}$, and concatenate them with a reflection prompt to form a reflection trajectory, \ie, $(s_{\text{neg}}, \text{prompt}_\text{reflect}, s)$.
We then use a function $\text{Replace}(\cdot)$ that replaces $s \in Y$ with $(s_{\text{neg}}, \text{prompt}_\text{reflect}, s)$ to convert $Y$ into the reflective reasoning path $Y_{\text{reflect}}$:
\begin{equation}
    Y_{\text{reflect}} = \text{Replace}(Y, s, (s_{\text{neg}}, \text{prompt}_\text{reflect}, s)),
\label{eq:reflect}
\end{equation}
where $\text{prompt}_\text{reflect}$ denotes a reflection prompt, such as ``The previous
reasoning step is wrong and let's rethink it again.''
Then, we can integrate the reflective reasoning path $Y_{\text{reflect}}$ into our data as a quadruplet $\{Q, Y, Y_{\text{reflect}}, S\} \in D$.

\normalem
\begin{algorithm}
\caption{Training Mulberry with CoMCTS}
\label{algo:comcts}
\begin{algorithmic}
    \STATE {\bfseries Input:} a set of policy models $\{\pi_{1}, \pi_{2},...,\pi_{K}\}$ initialized by different MLLMs; a set of multimodal questions $D_{Q}$
    \FOR{\textit{i = 1 to MaxEpoch}}
    \STATE \textcolor{gray}{Reasoning Tree Search using CoMCTS:}
    \FOR{$Q \in D_{Q}$}
    \STATE \textcolor{gray}{Collective Monte Carlo tree search:}
    \STATE $\{Q, Y, S\} = \text{CoMCTS}(\{\pi_{1}, \pi_{2},...,\pi_{K}\};Q)$ 
    \IF{\textit{found an effective reasoning path}}{
     \STATE Search and find $Y_{\text{reflect}}$ from $S$
     \STATE Add $\{Q, Y, Y_{\text{reflect}}, S\}$ into $D$
     \STATE Remove $Q$ from $D_{Q}$}
    \ENDIF
    \ENDFOR
    \STATE \textcolor{gray}{Model Training with CoMCTS Reasoning Trees:}
    \FOR{\textit{k = 1 to K}}
    \FOR{$(Q, Y, Y_{\text{reflect}}, S) \in D$}
    \STATE \textcolor{gray}{Supervised Fine-Tuning:}
    \STATE Optimize $\pi_{k}$ via $\mathcal{L}_{\text{CoSFT}}(\pi_{k})$ and $\mathcal{L}_{\text{CoSFT-Re}}(\pi_{k})$ 
  \ENDFOR
  \ENDFOR
  \ENDFOR
  \STATE {\bfseries Output:} {Trained policy models $\{\pi_{1}, \pi_{2},...,\pi_{K}\}$}
\end{algorithmic}
\end{algorithm}
\ULforem

\subsection{Training with Collective MCTS}

Using CoMCTS, we search effective and reflective reasoning paths for a set of multimodal input questions, and construct Mulberry-260k, a multimodal learning-to-reason-and-reflect dataset with a tree of rich, explicit and well-defined reasoning nodes for each question, \ie, a set of quadruplets $\{Q, Y, Y_{\text{reflect}}, S\} \in D$.
To learn collective knowledge from Mulberry-260k, we perform collective SFT to train our model, Mulberry, a series of Multimodal LLMs with o1-like step-by-step Reasoning and Reflection capabilities.

\begin{figure*}[ht]
\centering
\includegraphics[width=1\linewidth]{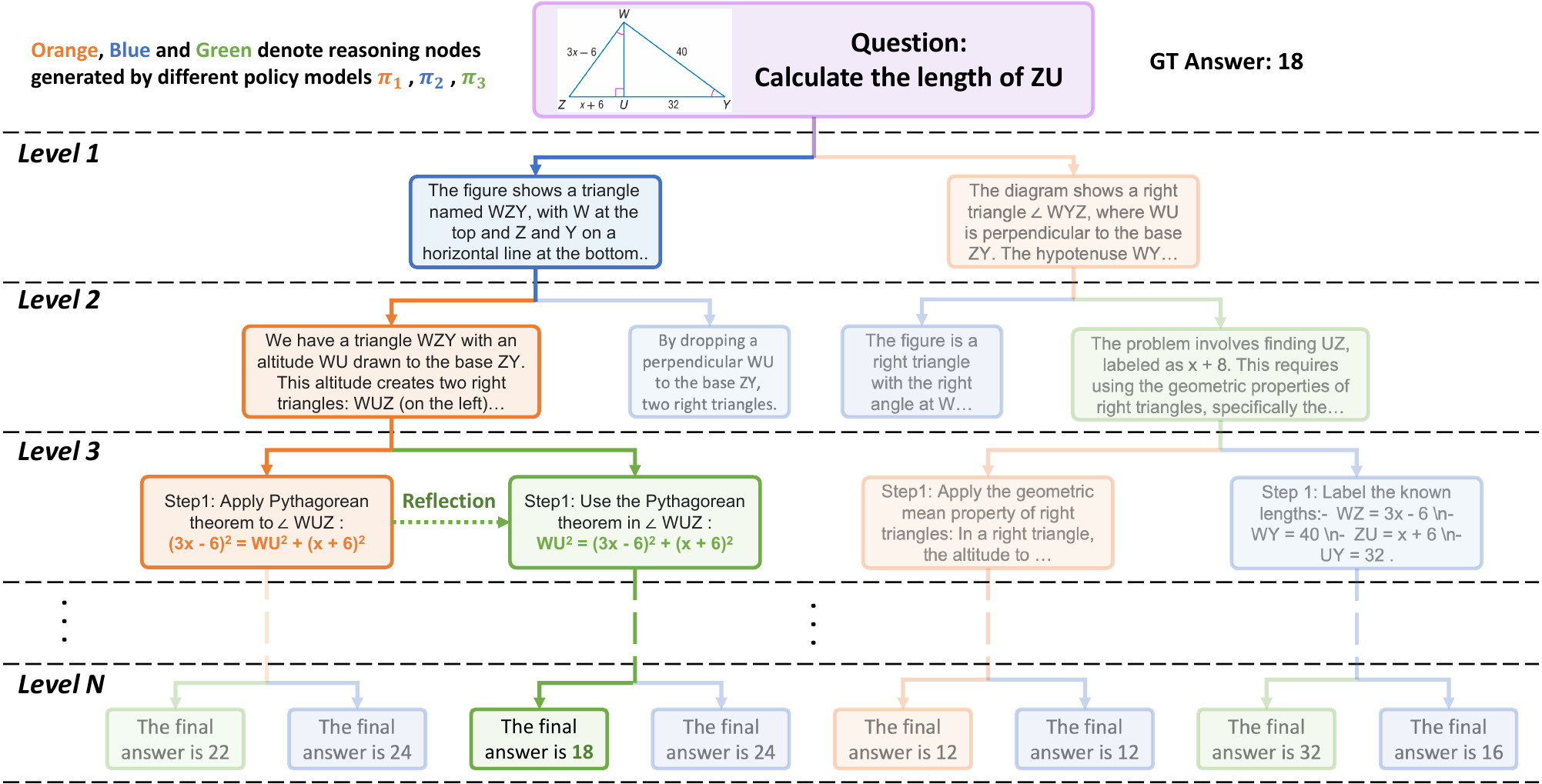}
\caption{Qualitative illustration of reasoning tree searched by CoMCTS with rich, explicit, well-defined reasoning nodes.}
\label{fig:vis schematic}
\vspace{-1.0em}
\end{figure*}

\textbf{Collective Supervised Fine-Tuning (CoSFT)}.
Given $(Q, Y) \in \mathcal{D}$, we apply standard SFT objective to train our MLLM to learn from $D$ constructed by CoMCTS:
\begin{equation}
\mathcal{L}_{\text{CoSFT}}(\pi_{k}) = \sum_{(Q, Y) \in \mathcal{D}} \log \pi_{k}(Y|Q),
\end{equation}
where $Y = \{s\}$ denotes the effective reasoning path that includes a sequence of reasoning nodes collectively conjectured, searched and identified by a group of MLLMs.

\textbf{CoSFT for reflective reasoning}. Given a question and its reasoning tree $(Q, S) \in \mathcal{D}$ constructed by CoMCTS, we randomly sample a reflective reasoning path $Y_{\text{reflect}}$ from $S$ as in Eqs.\ref{eq:neg_sibling}-\ref{eq:reflect}, and conduct CoSFT for reflective reasoning:
\begin{equation}
\mathcal{L}_{\text{CoSFT-Re}}(\pi_{k}) = \sum_{(Q, Y_{\text{reflect}}) \in \mathcal{D}} \log \pi_{k}(Y_{\text{reflect}}|Q),
\end{equation}
where $Y_{\text{reflect}} = \{s\}$ denotes the reflective reasoning path that includes an additional step-wise reflection trajectory.

The goal of $\mathcal{L}_{\text{CoSFT}}$ and $\mathcal{L}_{\text{CoSFT-Re}}$ is to maximize the log probability of effective and reflective reasoning path $Y$ and $Y_{\text{reflect}}$ over a tree of reasoning nodes $S$ generated by CoMCTS. 
In addition, $\mathcal{L}_{\text{CoSFT-Re}}$ enables to leverage the negative information during CoMCTS search process by learning to calibrate negative reasoning nodes.

\section{Experiment}
In this section, we first introduce our CoMCTS-generated dataset, Mulberry-260K, including its sources, construction, and analysis in \cref{subsec:dataset}, and provide implementation details in \cref{subsec:Implementation Details}.
We then present the main results in \cref{sec:exp_result}, demonstrating the effectiveness of the searched data (\ie, Mulberry-260K) and the trained models (\ie, Mulberry).
In \cref{sec:exp_ablation}, we perform comprehensive ablation studies on the impact of effective and reflective reasoning data and the contributions of collective knowledge sources.
In final, \cref{sec:exp_discussion} discuses the effectiveness and efficiency of tree search methods, explores different training strategies, and provides qualitative comparisons.

\begin{figure*}[t]
\centering
\includegraphics[width=0.29\linewidth]{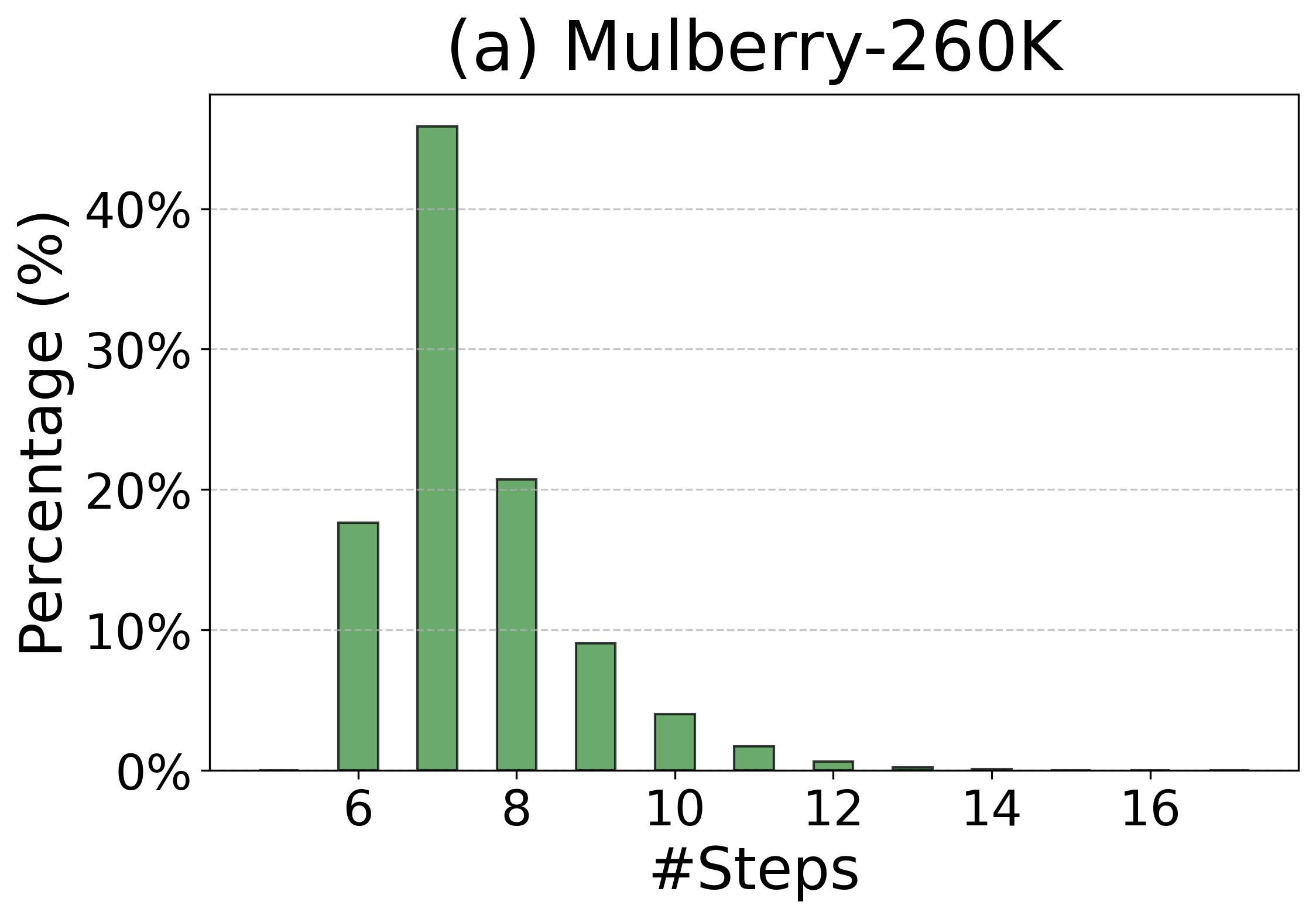}
\includegraphics[width=0.29\linewidth]{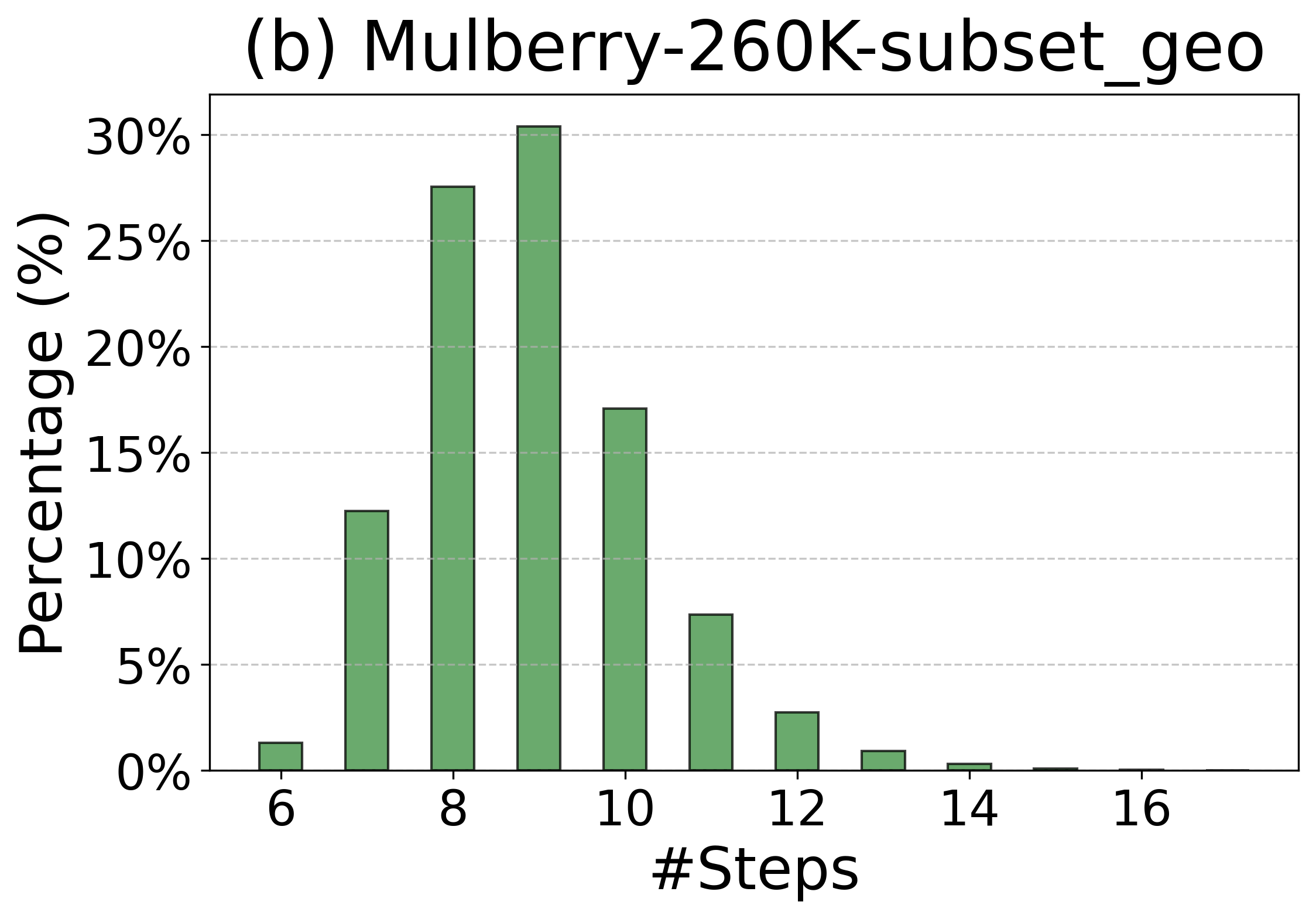}
\includegraphics[width=0.29\linewidth]{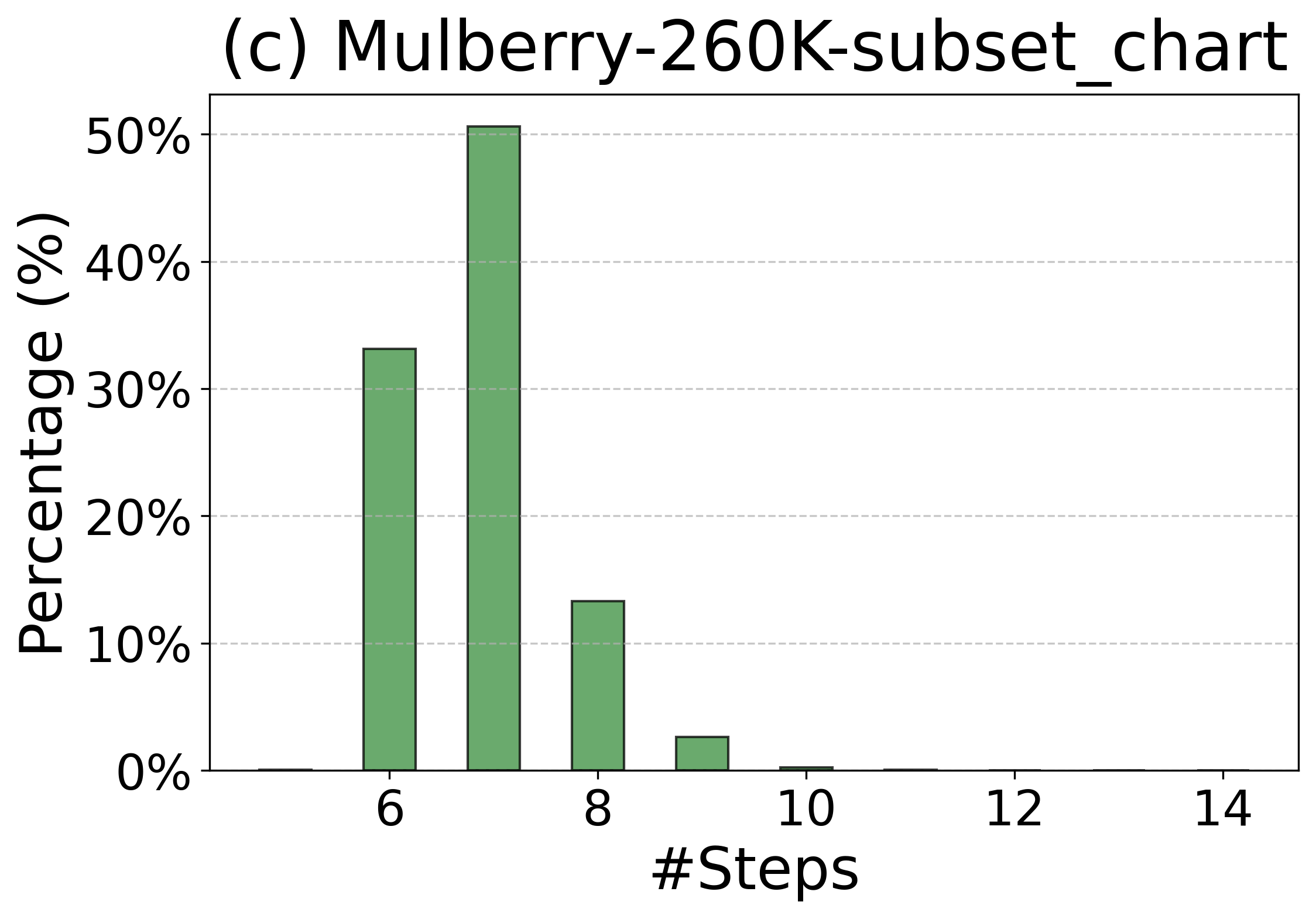}
\caption{
Distribution of reasoning steps in Mulberry-260K data.
} 
\label{fig:Step Distribution} 
\vspace{-1.0em}
\end{figure*}

\subsection{Dataset}
\label{subsec:dataset}

\textbf{The Sources of Raw Data.} To construct a comprehensive and general-purpose tree-based reasoning dataset, we collect 260K raw multimodal input questions (\ie, a text task instruction with an image as an input question) from a wide range of domains, covering General Multimodal Understanding, Mathematics, Figure Understanding, Realworld Understanding, Science, Medical Image Understanding, etc. The specific data sources are provided in the \cref{app: data sources}.

\textbf{Reasoning Data Construction.} As detailed in \cref{sec:method} and \cref{algo:comcts} and visually illustrated in \cref{fig:schematic,fig:vis schematic}, we employ our CoMCTS to search effective and reflective reasoning paths for a set of raw multimodal input questions as collected from the mentioned ``The Sources of Raw Data'', ultimately constructing our dataset, Mulberry-260K. Note we only sample 15K data for reflective reasoning training to avoid overabundance of reflection data.

\textbf{Reasoning Data Distribution.}
We analyze the CoMCTS-searched reasoning paths in Mulberry-260K by examining the distribution of reasoning steps, as shown in \cref{fig:Step Distribution}.
Specifically, \cref{fig:Step Distribution} shows that reasoning steps predominantly falls between 6 and 8, with an average of 7.5, for the entire Mulberry-260k.
Meanwhile, for simple reasoning tasks, the chart-related subset of Mulberry-260k, reasoning steps typically ranges from 6 to 7, averaging 6.8. For complex mathematical and logical reasoning tasks, such as the geometry-related subset of Mulberry-260k, the distribution shifts and largely falls between 7 and 10 steps, with an average of 8.9.
These observations highlight that the collective tree search design in CoMCTS enables to generate effective reasoning trajectories with flexible numbers of reasoning steps, learning from which allows to train a powerful MLLM with great reasoning flexibility, \ie, a model can ``think less and faster'' when handling simple questions (\ie, allocate and generate fewer intermediate reasoning steps) and ``think more and slower'' when tackling complex tasks (\ie, allocate and generate a greater number of intermediate reasoning steps).

\begin{table*}[ht]
  \centering
  \scalebox{0.85}{
  \setlength{\tabcolsep}{2pt}
  \begin{tabular}{@{}lccccccccl@{}}
    \toprule
    Method & MathVista & MMStar & MMMU  & ChartQA & DynaMath & HallBench & MM-Math & MME$_{sum}$ & \makecell[c]{AVG} \\
    \midrule
    \textit{Closed-Source Model} \\
    GPT-4o~\cite{gpt4o} & 63.8 & 63.9 & 69.1 & 85.7 & 63.7 & 55.0 & 31.8 & 2329 & 64.5 \\
    Claude-3.5 Sonnet~\cite{claude_3.5_sonnet}  & 67.7 & 62.2 & 68.3 & 90.8 & 64.8 & 55.0 & - & 1920 & \makecell[c]{-} \\
    \midrule
    \textit{Open-Source Model} & & & &\\
    DeepSeek-VL-7B~\cite{deepseek-vl} & 36.1 & 37.1 & 35.4 & 59.1 & 21.5 & - & - & - & \makecell[c]{-}\\ 
    Cambrain-1-8B~\cite{Cambrian-1} & 49.0 & - & 42.7 & 73.3 & - & - & - & - &  \makecell[c]{-}\\ 
    MM-1.5-7B~\cite{mm1.5} & 47.6 & - & 41.8 & 78.6 & - & - & - & 1861 & \makecell[c]{-} \\ 
    Idefics3-LLaMA3-8B~\cite{idefics3} & 58.4 & 55.9 & 46.6 & 74.8 & - & - & - & 1937 & \makecell[c]{-}\\ 
    InternVL2-8B~\cite{internvl2} & 58.3 & \textbf{61.5} & 51.8 & 83.3 & 39.7 & - & - & 2210 & \makecell[c]{-}\\
    MiniCPM-Llama-V-2.5-8B~\cite{minicpm-v} & 54.3 & 51.8 & 45.8 & - & - & 42.4 & - & 2025  & \makecell[c]{-}\\ 
    MiniCPM-V-2.6-8B~\cite{minicpm-v} & 60.6 & 57.5 & 49.8 & - & - & 48.1 & - & 2348 & \makecell[c]{-} \\
    DeepSeek-VL2-MOE-4.5B~\cite{deepseek-vl2} & 62.8 & 61.3 & 51.1 & 86.0 & - & - & - & 2253 & \makecell[c]{-} \\
    \midrule
    \textit{Reasoning Model} & & & &\\ 
    LLaVA-CoT-11B~\cite{llava-cot} & 54.8 & 57.6 & - & - & - & 47.8 & - & - & \makecell[c]{-}\\ 
    LLaVA-Reasoner-8B~\cite{llava-reasoner} & 50.6 & 54.0 & 40.0 & 83.0 & - & - & - & - & \makecell[c]{-} \\ 
    Insight-V-8B~\cite{insight-v} & 49.8 & 57.4 & 42.0 & 77.4 & - & - & - & 2069 & \makecell[c]{-}\\ 
    \midrule
    LLaVA-NeXT-8B~\cite{llavanext-strong} & 37.5 & 42.1 & 41.7  & 69.5 & 22.7 & 33.4 & 0.6 & 1957 & 39.7\\
    \rowcolor{mygray}
    Mulberry-LLaVA-8B & 56.3 &  54.5 & 43.0  & 79.5 & 34.1 & 47.5 & 18.9 & 2021 & 50.7$^{\color{teal}{\textbf{\scriptsize11}\uparrow}}$\\
    Llama-3.2-11B-V-Ins.~\cite{llama3} & 48.6 & 49.8 & 41.7 & 83.4 & 34.3 & 40.3 & 4.1 & 1787 & 45.8 \\
    \rowcolor{mygray}
    Mulberry-Llama-11B & 61.1 & 58.5 & 45.6 & 83.5 & 37.2 & 48.9 & 18.7 & 2035 & 53.3$^{\color{teal}{\textbf{\scriptsize7.5}\uparrow}}$ \\ 
    \midrule
    Qwen2-VL-2B~\cite{qwen2vl} & 43.0 & 48.0 & 41.1 & 73.5 & 24.9 & 41.7 & 1.0  & 1872 & 42.5\\ 
    \rowcolor{mygray}
    Mulberry-2B & 51.7 & 51.3 & 42.0  & 77.7 & 30.0 & 44.9 & 13.9 & 2013 & 47.9$^{\color{teal}{\textbf{\scriptsize5.4}\uparrow}}$\\
    Qwen2-VL-7B~\cite{qwen2vl} & 58.2 & 60.7 & 54.1 & 83.0 & 42.1 & 50.6 & 5.9 &2327 & 54.7\\
    \rowcolor{mygray}
    Mulberry-7B  & \textbf{63.1} & 61.3 & \textbf{55.0} & \textbf{83.9} & \textbf{45.1} & \textbf{54.1} & \textbf{23.7} & \textbf{2396} & \textbf{58.9}$^{\color{teal}{\textbf{\scriptsize4.2}\uparrow}}$ \\
    \bottomrule
  \end{tabular}}
  \caption{\textbf{Main Results.} To examine the effectiveness of the searched data (\ie, Mulberry-260K) and the trained models (\ie, Mulberry), we conduct extensive experiments with four powerful baseline models, and comprehensively benchmark our Mulberry with various state-of-the-arts, including general and reasoning-based MLLMs. }
  \label{tab:Main Results}
  \vspace{-1.0em}
\end{table*}

\begin{table}[ht]
  \centering
  \resizebox{0.99\linewidth}{!}{
  \setlength{\tabcolsep}{3pt}
  \begin{tabular}{@{}c|cccc|c@{}}
    \toprule
    Direct Pred &\multicolumn{4}{c|}{CoMCTS} & \multirow{2.5}{*}{S.S.R.} \\
    \cmidrule(){1-5}
    GPT-4o &GPT-4o & Qwen2-VL-7B & LLama3.2-11B & Qwen2-VL-72B & ~ \\
    \midrule
    \CheckmarkBold & & & & & 58.2 \\
    &\CheckmarkBold & & & & 63.8 \\
    &\CheckmarkBold & \CheckmarkBold & & & 66.2 \\
    &\CheckmarkBold & \CheckmarkBold & \CheckmarkBold & & 69.7 \\
    &\CheckmarkBold & \CheckmarkBold & \CheckmarkBold & \CheckmarkBold & 80.2 \\
    \bottomrule
  \end{tabular}}
  \caption{
  \textbf{Ablation Study on CoMCTS.} 
  We study how each model in CoMCTS collective learning contribute to overall tree search performance in Search Success Rate (S.S.R.).
  }
  \label{tab:Ablation Study on the Search Models.}
\end{table}

\subsection{Implementation Detail}
\label{subsec:Implementation Details}
In this paper, we implement the collective learning in CoMCTS by employing a group of four models, including GPT-4o, Qwen2-VL-7B, LLaMA-3.2-11B-Vision-Instruct, and Qwen2-VL-72B, to construct Mulberry-260K. 
In our CoMCTS, we set the maximum search iteration as $20$. In each search iteration, we employ each model from the group to generate one subsequent candidate reasoning path to balance search exploration and exploitation. In Simulation and Error Positioning in CoMCTS, we simply set threshold $t$ as 0. 
We adopt four popular MLLMs as baseline models, and conduct experiments on baselines Qwen2-VL-7B and LLaMA-3.2-11B-Vision-Instruct to examine the search effectiveness of our CoMCTS, and on baselines Qwen2-VL-2B and LLaVA-NeXT-8B to study the generalization of CoMCTS-searched data.
The collective SFT experiments are conducted with a batch size of 128, a learning rate of 1e-5, and training over 2 epochs. For Qwen2-VL-7B, a smaller learning rate of 5e-6 is adopted to stabilize the training.

\subsection{Main Results}
\label{sec:exp_result}
To examine the effectiveness of the searched data (\ie, Mulberry-260K) and the trained models (\ie, Mulberry), we conduct extensive experiments with four powerful baseline models, and comprehensively benchmark our Mulberry with various state-of-the-arts, including general and reasoning-based MLLMs.
The evaluation is performed on 8 widely used and challenging datasets~\cite{evaluation_survey}, covering the fields ranging from general and mathematical reasoning to hallucination and visual illusion, and multi-disciplinary understanding and reasoning, as shown in \cref{tab:Main Results}.

\textbf{Comparison with baselines.}
We first conduct experiments on baselines Qwen2-VL-7B and LLaMA-3.2-11B-Vision-Instruct that are involved in collective learning of CoMCTS for joint reasoning-path conjecture, search and identification.
We can observe that, trained with jointly-searched data (\ie, Mulberry-260k), our Mulberry-7B and Mulberry-11B bring clear performance improvements against their baselines, \ie, +4.2\% over Qwen2-VL-7B and +7.5\% over LLaMA-3.2-11B-Vision-Instruct averaged on 8 benchmarks, validating the search effectiveness of our CoMCTS.
On the other hand, we examine the generalization of our Mulberry-260k by applying it to train other models that are not involved in collective tree search in CoMCTS, such as Qwen2-VL-2B and LLaVA-NeXT-8B.
It can be observed that, trained with Mulberry-260k, our models (\ie, Mulberry-2B and Mulberry-8B) enhance Qwen2-VL-2B and LLaVA-NeXT-8B with +5.4\% and +11.0\% gains averaged on 8 benchmarks, demonstrating the generalization of CoMCTS-searched data.

\textbf{Comparison with reasoning-response models.}
We then benchmark our Mulberry with various state-of-the-art reasoning-response models.
It shows that, using the same base model LLaVA-NeXT-8B~\cite{llavanext-strong}, our Mulberry outperforms LLaVA-Reasoner-8B and Insight-V-8B by +5.7\% and +6.5\% on mathematical benchmark MathVista, and by +3.0\% and +1.0\% on multi-disciplinary benchmark MMMU, respectively.
Besides, Mulberry-11B surpasses LLaVA-COT-11B by +6.3\% on reasoning-intensive benchmark MathVista under the same baseline LLaMA-3.2-11B-Vision-Instruct.
The great superiority of Mulberry is largely attributed to our CoMCTS that conducts tree search and provides rich, explicit and well-defined reasoning nodes with flexible numbers of steps. 

\textbf{Comparison with state-of-the-arts.} 
In final, we benchmark our Mulberry with popular state-of-the-arts included both open-source and closed-source ones.
The results in \cref{tab:Main Results} show that our Mulberry, trained on CoMCTS-searched data, outperforms most open-sourced MLLMs and achieves competitive results against closed-source ones, demonstrating outstanding abilities in step-by-step reasoning and reflection.

\begin{table}[t]
  \centering
  \resizebox{0.99\linewidth}{!}{
  \setlength{\tabcolsep}{13.0pt}
  \begin{tabular}{@{}lcc@{}}
    \toprule
    Benchmark & w/o Reflection Data & w/ Reflection Data \\
    \midrule
    MathVista & 50.9 & 51.7 \\
    \bottomrule
  \end{tabular}}
  \caption{
  \textbf{Ablation Study on Mulberry.} As Mulberry is trained with effective and reflective reasoning data searched by CoMCTS, we study their respective contributions.
  }
  \label{tab: Ablation Study on Reflection Data.}
\end{table}

\begin{table}[t]
    \centering
    \scalebox{0.8}{
    \setlength{\tabcolsep}{5.0pt}
    \begin{tabular}{@{}l|cccc@{}}
    \toprule
    Methods & Search Success Rate~\textbf{$\uparrow$} & Average Search Iteration~\textbf{$\downarrow$}\\
    \midrule
    GPT4o (direct) & 58.2 & - \\
    MCTS & 63.8 & 42.1 \\
    ReST-MCTS & 65.6 & 36.3\\
    Omega-MCTS & 66.2 & 24.3\\
    CoMCTS & 80.2 & 12.7 \\
    \bottomrule
  \end{tabular}
  }
  \caption{\textbf{Comparison with other tree search methods.} 
``GPT-4o (direct)'' refers to the baseline without tree search.
  Our CoMCTS shows great superiority in search effectiveness and efficiency.
  }
  \label{tab:search methods comparison}
\end{table}

\subsection{Ablation Study}\label{sec:exp_ablation}
\textbf{Ablation Study on CoMCTS.} 
We conduct ablation studies with the powerful GPT-4o as the baseline over 1K samples from Geo3K~\cite{geo3k} and GeoQA-Plus~\cite{chen-etal-2021-geoqa}, as shown in~\cref{tab:Ablation Study on the Search Models.}.
As the core of our proposed CoMCTS, we examine how each model in the collective learning group contribute to the overall tree search performance. 
\cref{tab:Ablation Study on the Search Models.} reports the Search Success Rates (S.S.R.). 
The baseline GPT-4o works not very well without tree search. 
It shows that CoMCTS with only GPT-4o improves the performance to 63.8\%, largely becuase our tree search designs like expansion, simulation and error positioning can work even without using collective knowledge.
Besides, progressively involving more models into CoMCTS consistently improves the search performance, even for including small models like Qwen2-VL-7B (\ie, +2.4\%), demonstrating the effectiveness of CoMCTS in capturing useful collective knowledge not only with large models but also from small models.
In final, the inclusion of all four models in the proposed CoMCTS performs clearly the best, \ie 80.2\%, validating the effectiveness of collective learning on reasoning tree search.

\begin{figure*}[t]
\centering
\includegraphics[width=0.87\linewidth]{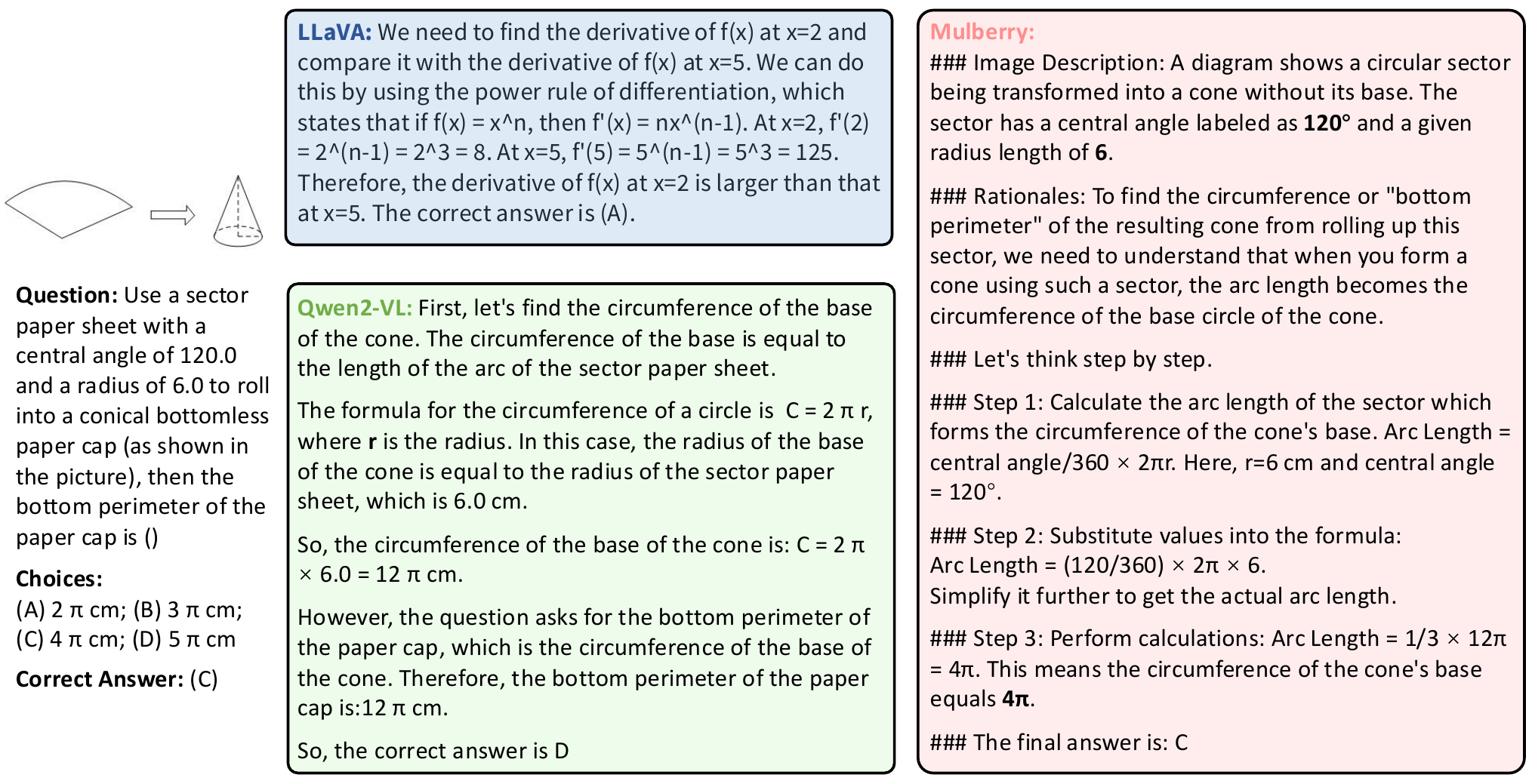}
\caption{Qualitative Comparison. Our Mulberry, trained with CoMCTS-searched reasoning data, creates rich, explicit and well-defined reasoning steps with comprehensive understanding, ultimately arriving at the correct answer.}
\label{fig:qualitative results}
\vspace{-1.0em}
\end{figure*}

\textbf{Ablation Study on Mulberry.} 
We train Mulberry with effective and reflective reasoning data searched by CoMCTS, and study their respective contributions to overall reasoning performance. 
\cref{tab: Ablation Study on Reflection Data.} presents the results on MathVista, which show that incorporating reflection data enhances the performance by 0.8\%, demonstrating the complementarity of effective and reflective reasoning data searched by CoMCTS.

\subsection{Discussion}\label{sec:exp_discussion}
\textbf{Comparison with other tree search methods.}
We compare our CoMCTS with other tree search methods in search effectiveness and efficiency, including the baseline ``GPT-4o direction prediction'', ``traditional MCTS~\cite{coulom2006efficient}'', ``ReST-MCTS~\cite{zhang2024rest}'' that enhances MCTS by introducing partial search, and ``Omega-MCTS~\cite{omegaprm}'' that improves MCTS by designing binary search. 
\cref{tab:search methods comparison} shows the results in search success rate and average search iteration that indicate search effectiveness and efficiency respectively.
We can observe that existing tree search methods improve GPT-4o with limited gains. One main reason lies in that traditional MCTS methods generally work by self-bootstrapping and often get trapped in homogeneous low-quality nodes within the reasoning space of a single MLLM.
On the other hand, our CoMCTS shows great superiority in search effectiveness and efficiency, largely thanks to the joint expansion mechanism in CoMCTS that allows reasoning-path search not only within the reasoning space of a given MLLM itself but also among those of others, benefiting from the synergy of multiple MLLMs while avoiding being trapped within the reasoning space of a single MLLM.

\textbf{Qualitative comparison.}
We provide qualitative comparison of LLaVA-NeXT-8B~\cite{llavanext-strong}, Qwen2-VL-7B~\cite{qwen2vl}, and Mulberry-7B in \cref{fig:qualitative results}.
It shows that LLaVA-NeXT-8B and Qwen2-VL-7B generate relatively short predictions without thorough thinking, leading to incorrect answers.
On the contrary, our Mulberry, trained with CoMCTS-searched reasoning data, creates rich, explicit and well-defined reasoning steps with comprehensive understanding, ultimately arriving at the correct answer.

\section{Conclusion}
This paper presents CoMCTS, a new learning-to-reason approach for MLLMs, which introduces the concept of collective learning into “tree search” for effective and efficient reasoning-path searching and learning.
Based on the proposed CoMCTS, we search effective and reflective reasoning paths for a set of multimodal inputs, and construct Mulberry-260k, a multimodal learning-to-reason-and-reflect dataset with a tree of rich, explicit and well-defined reasoning nodes for each question. 
Using Mulberry-260k, we train our model, Mulberry, a series of Multimodal LLMs with o1-like step-by-step Reasoning and Reflection capabilities.
Furthermore, we conduct extensive experiments, ablation studies and discussion, which demonstrate the superiority of our proposed methods on various benchmarks.
We hope that CoMCTS along with Mulberry-260k and Mulberry will provides valuable resources and offer new insights for multimodal MCTS search and reasoning.

\clearpage
\nocite{rlhf-V}
\nocite{ovis}
\normalem
\clearpage
\bibliography{example_paper}

\begin{thebibliography}{80}
\providecommand{\natexlab}[1]{#1}
\providecommand{\url}[1]{\texttt{#1}}
\expandafter\ifx\csname urlstyle\endcsname\relax
  \providecommand{\doi}[1]{doi: #1}\else
  \providecommand{\doi}{doi: \begingroup \urlstyle{rm}\Url}\fi

\bibitem[Anthropic(2024)]{claude_3.5_sonnet}
Anthropic.
\newblock Claude 3.5 sonnet, 2024.
\newblock URL \url{https://www.anthropic.com/news/claude-3-5-sonnet}.

\bibitem[Antol et~al.(2015)Antol, Agrawal, Lu, Mitchell, Batra, Zitnick, and Parikh]{vqa-as}
Antol, S., Agrawal, A., Lu, J., Mitchell, M., Batra, D., Zitnick, C.~L., and Parikh, D.
\newblock Vqa: Visual question answering.
\newblock In \emph{Proceedings of the IEEE international conference on computer vision}, pp.\  2425--2433, 2015.

\bibitem[Best et~al.(2019)Best, Cliff, Patten, Mettu, and Fitch]{best2019dec}
Best, G., Cliff, O.~M., Patten, T., Mettu, R.~R., and Fitch, R.
\newblock Dec-mcts: Decentralized planning for multi-robot active perception.
\newblock \emph{The International Journal of Robotics Research}, 38\penalty0 (2-3):\penalty0 316--337, 2019.

\bibitem[Besta et~al.(2024)Besta, Blach, Kubicek, Gerstenberger, Podstawski, Gianinazzi, Gajda, Lehmann, Niewiadomski, Nyczyk, et~al.]{besta2024graph}
Besta, M., Blach, N., Kubicek, A., Gerstenberger, R., Podstawski, M., Gianinazzi, L., Gajda, J., Lehmann, T., Niewiadomski, H., Nyczyk, P., et~al.
\newblock Graph of thoughts: Solving elaborate problems with large language models.
\newblock In \emph{Proceedings of the AAAI Conference on Artificial Intelligence}, volume~38, pp.\  17682--17690, 2024.

\bibitem[Blum \& Mitchell(1998)Blum and Mitchell]{blum1998combining}
Blum, A. and Mitchell, T.
\newblock Combining labeled and unlabeled data with co-training.
\newblock In \emph{Proceedings of the eleventh annual conference on Computational learning theory}, pp.\  92--100, 1998.

\bibitem[Chen et~al.(2021)Chen, Tang, Qin, Liang, Liu, Xing, and Lin]{chen-etal-2021-geoqa}
Chen, J., Tang, J., Qin, J., Liang, X., Liu, L., Xing, E.~P., and Lin, L.
\newblock Geoqa: A geometric question answering benchmark towards multimodal numerical reasoning.
\newblock \emph{arXiv preprint arXiv:2105.14517}, 2021.

\bibitem[Chen et~al.(2022)Chen, Li, Qin, Lu, Lin, Chen, and Liang]{chen2022unigeo}
Chen, J., Li, T., Qin, J., Lu, P., Lin, L., Chen, C., and Liang, X.
\newblock Unigeo: Unifying geometry logical reasoning via reformulating mathematical expression.
\newblock \emph{arXiv preprint arXiv:2212.02746}, 2022.

\bibitem[Chen et~al.(2024)Chen, Wang, Tian, Ye, Gao, Cui, Tong, Hu, Luo, Ma, et~al.]{internvl2}
Chen, Z., Wang, W., Tian, H., Ye, S., Gao, Z., Cui, E., Tong, W., Hu, K., Luo, J., Ma, Z., et~al.
\newblock How far are we to gpt-4v? closing the gap to commercial multimodal models with open-source suites.
\newblock \emph{arXiv preprint arXiv:2404.16821}, 2024.

\bibitem[Coulom(2006)]{coulom2006efficient}
Coulom, R.
\newblock Efficient selectivity and backup operators in monte-carlo tree search.
\newblock In \emph{International conference on computers and games}, pp.\  72--83. Springer, 2006.

\bibitem[Cui et~al.(2022)Cui, Huang, Luo, Zhang, Zhan, and Lu]{cui2022genco}
Cui, K., Huang, J., Luo, Z., Zhang, G., Zhan, F., and Lu, S.
\newblock Genco: Generative co-training for generative adversarial networks with limited data.
\newblock In \emph{Proceedings of the AAAI Conference on Artificial Intelligence}, volume~36, pp.\  499--507, 2022.

\bibitem[Dam et~al.(2022)Dam, Chalvatzaki, Peters, and Pajarinen]{dam2022monte}
Dam, T., Chalvatzaki, G., Peters, J., and Pajarinen, J.
\newblock Monte-carlo robot path planning.
\newblock \emph{IEEE Robotics and Automation Letters}, 7\penalty0 (4):\penalty0 11213--11220, 2022.

\bibitem[Dong et~al.(2024)Dong, Liu, Sun, Yang, Hu, Rao, and Liu]{insight-v}
Dong, Y., Liu, Z., Sun, H.-L., Yang, J., Hu, W., Rao, Y., and Liu, Z.
\newblock Insight-v: Exploring long-chain visual reasoning with multimodal large language models.
\newblock \emph{arXiv preprint arXiv:2411.14432}, 2024.

\bibitem[Dubey et~al.(2024)Dubey, Jauhri, Pandey, Kadian, Al-Dahle, Letman, Mathur, Schelten, Yang, Fan, et~al.]{llama3}
Dubey, A., Jauhri, A., Pandey, A., Kadian, A., Al-Dahle, A., Letman, A., Mathur, A., Schelten, A., Yang, A., Fan, A., et~al.
\newblock The llama 3 herd of models.
\newblock \emph{arXiv preprint arXiv:2407.21783}, 2024.

\bibitem[Fawzi et~al.(2022)Fawzi, Balog, Huang, Hubert, Romera-Paredes, Barekatain, Novikov, R~Ruiz, Schrittwieser, Swirszcz, et~al.]{fawzi2022discovering}
Fawzi, A., Balog, M., Huang, A., Hubert, T., Romera-Paredes, B., Barekatain, M., Novikov, A., R~Ruiz, F.~J., Schrittwieser, J., Swirszcz, G., et~al.
\newblock Discovering faster matrix multiplication algorithms with reinforcement learning.
\newblock \emph{Nature}, 610\penalty0 (7930):\penalty0 47--53, 2022.

\bibitem[Foerster et~al.(2016)Foerster, Assael, De~Freitas, and Whiteson]{foerster2016learning}
Foerster, J., Assael, I.~A., De~Freitas, N., and Whiteson, S.
\newblock Learning to communicate with deep multi-agent reinforcement learning.
\newblock \emph{Advances in neural information processing systems}, 29, 2016.

\bibitem[Gao et~al.(2023)Gao, Pi, Zhang, Ye, Zhong, Wang, Hong, Han, Xu, Li, et~al.]{g-llava}
Gao, J., Pi, R., Zhang, J., Ye, J., Zhong, W., Wang, Y., Hong, L., Han, J., Xu, H., Li, Z., et~al.
\newblock G-llava: Solving geometric problem with multi-modal large language model.
\newblock \emph{arXiv preprint arXiv:2312.11370}, 2023.

\bibitem[Goyal et~al.(2017)Goyal, Khot, Summers-Stay, Batra, and Parikh]{vqa2.0}
Goyal, Y., Khot, T., Summers-Stay, D., Batra, D., and Parikh, D.
\newblock Making the v in vqa matter: Elevating the role of image understanding in visual question answering.
\newblock In \emph{Proceedings of the IEEE conference on computer vision and pattern recognition}, pp.\  6904--6913, 2017.

\bibitem[Gurari et~al.(2018)Gurari, Li, Stangl, Guo, Lin, Grauman, Luo, and Bigham]{gurari2018vizwiz}
Gurari, D., Li, Q., Stangl, A.~J., Guo, A., Lin, C., Grauman, K., Luo, J., and Bigham, J.~P.
\newblock Vizwiz grand challenge: Answering visual questions from blind people.
\newblock In \emph{Proceedings of the IEEE conference on computer vision and pattern recognition}, pp.\  3608--3617, 2018.

\bibitem[Huang \& Zhang(2024)Huang and Zhang]{evaluation_survey}
Huang, J. and Zhang, J.
\newblock A survey on evaluation of multimodal large language models.
\newblock \emph{arXiv preprint arXiv:2408.15769}, 2024.

\bibitem[Hurst et~al.(2024)Hurst, Lerer, Goucher, Perelman, Ramesh, Clark, Ostrow, Welihinda, Hayes, Radford, et~al.]{gpt4o}
Hurst, A., Lerer, A., Goucher, A.~P., Perelman, A., Ramesh, A., Clark, A., Ostrow, A., Welihinda, A., Hayes, A., Radford, A., et~al.
\newblock Gpt-4o system card.
\newblock \emph{arXiv preprint arXiv:2410.21276}, 2024.

\bibitem[Johnson et~al.(2017)Johnson, Hariharan, Van Der~Maaten, Fei-Fei, Lawrence~Zitnick, and Girshick]{johnson2017clevr}
Johnson, J., Hariharan, B., Van Der~Maaten, L., Fei-Fei, L., Lawrence~Zitnick, C., and Girshick, R.
\newblock Clevr: A diagnostic dataset for compositional language and elementary visual reasoning.
\newblock In \emph{Proceedings of the IEEE conference on computer vision and pattern recognition}, pp.\  2901--2910, 2017.

\bibitem[Kafle et~al.(2018)Kafle, Price, Cohen, and Kanan]{kafle2018dvqa}
Kafle, K., Price, B., Cohen, S., and Kanan, C.
\newblock Dvqa: Understanding data visualizations via question answering.
\newblock In \emph{Proceedings of the IEEE conference on computer vision and pattern recognition}, pp.\  5648--5656, 2018.

\bibitem[Kahou et~al.(2017)Kahou, Michalski, Atkinson, K{\'a}d{\'a}r, Trischler, and Bengio]{kahou2017figureqa}
Kahou, S.~E., Michalski, V., Atkinson, A., K{\'a}d{\'a}r, {\'A}., Trischler, A., and Bengio, Y.
\newblock Figureqa: An annotated figure dataset for visual reasoning.
\newblock \emph{arXiv preprint arXiv:1710.07300}, 2017.

\bibitem[Kazemi et~al.(2023)Kazemi, Alvari, Anand, Wu, Chen, and Soricut]{geomverse}
Kazemi, M., Alvari, H., Anand, A., Wu, J., Chen, X., and Soricut, R.
\newblock Geomverse: A systematic evaluation of large models for geometric reasoning.
\newblock \emph{arXiv preprint arXiv:2312.12241}, 2023.

\bibitem[Kembhavi et~al.(2016)Kembhavi, Salvato, Kolve, Seo, Hajishirzi, and Farhadi]{ai2d}
Kembhavi, A., Salvato, M., Kolve, E., Seo, M., Hajishirzi, H., and Farhadi, A.
\newblock A diagram is worth a dozen images.
\newblock In \emph{Computer Vision--ECCV 2016: 14th European Conference, Amsterdam, The Netherlands, October 11--14, 2016, Proceedings, Part IV 14}, pp.\  235--251. Springer, 2016.

\bibitem[Kembhavi et~al.(2017)Kembhavi, Seo, Schwenk, Choi, Farhadi, and Hajishirzi]{tqa}
Kembhavi, A., Seo, M., Schwenk, D., Choi, J., Farhadi, A., and Hajishirzi, H.
\newblock Are you smarter than a sixth grader? textbook question answering for multimodal machine comprehension.
\newblock In \emph{Proceedings of the IEEE Conference on Computer Vision and Pattern recognition}, pp.\  4999--5007, 2017.

\bibitem[Lample et~al.(2022)Lample, Lacroix, Lachaux, Rodriguez, Hayat, Lavril, Ebner, and Martinet]{lample2022hypertree}
Lample, G., Lacroix, T., Lachaux, M.-A., Rodriguez, A., Hayat, A., Lavril, T., Ebner, G., and Martinet, X.
\newblock Hypertree proof search for neural theorem proving.
\newblock \emph{Advances in neural information processing systems}, 35:\penalty0 26337--26349, 2022.

\bibitem[Lau et~al.(2018)Lau, Gayen, Ben~Abacha, and Demner-Fushman]{VQA-RAD}
Lau, J.~J., Gayen, S., Ben~Abacha, A., and Demner-Fushman, D.
\newblock A dataset of clinically generated visual questions and answers about radiology images.
\newblock \emph{Scientific data}, 5\penalty0 (1):\penalty0 1--10, 2018.

\bibitem[Lauren{\c{c}}on et~al.(2024)Lauren{\c{c}}on, Marafioti, Sanh, and Tronchon]{idefics3}
Lauren{\c{c}}on, H., Marafioti, A., Sanh, V., and Tronchon, L.
\newblock Building and better understanding vision-language models: insights and future directions.
\newblock In \emph{Workshop on Responsibly Building the Next Generation of Multimodal Foundational Models}, 2024.

\bibitem[Li et~al.(2024)Li, Zhang, Zhang, Guo, Zhang, Li, Zhang, Liu, and Li]{llavanext-strong}
Li, B., Zhang, K., Zhang, H., Guo, D., Zhang, R., Li, F., Zhang, Y., Liu, Z., and Li, C.
\newblock Llava-next: Stronger llms supercharge multimodal capabilities in the wild, May 2024.
\newblock URL \url{https://llava-vl.github.io/blog/2024-05-10-llava-next-stronger-llms/}.

\bibitem[Li et~al.(2023)Li, Wang, Stengel-Eskin, Kortylewski, Ma, Van~Durme, and Yuille]{super-clevr}
Li, Z., Wang, X., Stengel-Eskin, E., Kortylewski, A., Ma, W., Van~Durme, B., and Yuille, A.~L.
\newblock Super-clevr: A virtual benchmark to diagnose domain robustness in visual reasoning.
\newblock In \emph{Proceedings of the IEEE/CVF Conference on Computer Vision and Pattern Recognition}, pp.\  14963--14973, 2023.

\bibitem[Lindstr{\"o}m \& Abraham(2022)Lindstr{\"o}m and Abraham]{clevr-math}
Lindstr{\"o}m, A.~D. and Abraham, S.~S.
\newblock Clevr-math: A dataset for compositional language, visual and mathematical reasoning.
\newblock \emph{arXiv preprint arXiv:2208.05358}, 2022.

\bibitem[Liu et~al.(2023)Liu, Lin, Li, Wang, Yacoob, and Wang]{lrv_chart}
Liu, F., Lin, K., Li, L., Wang, J., Yacoob, Y., and Wang, L.
\newblock Mitigating hallucination in large multi-modal models via robust instruction tuning.
\newblock In \emph{The Twelfth International Conference on Learning Representations}, 2023.

\bibitem[Liu et~al.(2024)Liu, Li, Wu, and Lee]{llava}
Liu, H., Li, C., Wu, Q., and Lee, Y.~J.
\newblock Visual instruction tuning.
\newblock \emph{Advances in neural information processing systems}, 36, 2024.

\bibitem[Lu et~al.(2024{\natexlab{a}})Lu, Liu, Zhang, Wang, Dong, Liu, Sun, Ren, Li, Yang, et~al.]{deepseek-vl}
Lu, H., Liu, W., Zhang, B., Wang, B., Dong, K., Liu, B., Sun, J., Ren, T., Li, Z., Yang, H., et~al.
\newblock Deepseek-vl: towards real-world vision-language understanding.
\newblock \emph{arXiv preprint arXiv:2403.05525}, 2024{\natexlab{a}}.

\bibitem[Lu et~al.(2021{\natexlab{a}})Lu, Gong, Jiang, Qiu, Huang, Liang, and Zhu]{geo3k}
Lu, P., Gong, R., Jiang, S., Qiu, L., Huang, S., Liang, X., and Zhu, S.-C.
\newblock Inter-gps: Interpretable geometry problem solving with formal language and symbolic reasoning.
\newblock In \emph{The 59th Annual Meeting of the Association for Computational Linguistics (ACL)}, 2021{\natexlab{a}}.

\bibitem[Lu et~al.(2021{\natexlab{b}})Lu, Qiu, Chen, Xia, Zhao, Zhang, Yu, Liang, and Zhu]{lu2021iconqa}
Lu, P., Qiu, L., Chen, J., Xia, T., Zhao, Y., Zhang, W., Yu, Z., Liang, X., and Zhu, S.-C.
\newblock Iconqa: A new benchmark for abstract diagram understanding and visual language reasoning.
\newblock \emph{arXiv preprint arXiv:2110.13214}, 2021{\natexlab{b}}.

\bibitem[Lu et~al.(2022{\natexlab{a}})Lu, Mishra, Xia, Qiu, Chang, Zhu, Tafjord, Clark, and Kalyan]{scienceqa}
Lu, P., Mishra, S., Xia, T., Qiu, L., Chang, K.-W., Zhu, S.-C., Tafjord, O., Clark, P., and Kalyan, A.
\newblock Learn to explain: Multimodal reasoning via thought chains for science question answering.
\newblock \emph{Advances in Neural Information Processing Systems}, 35:\penalty0 2507--2521, 2022{\natexlab{a}}.

\bibitem[Lu et~al.(2022{\natexlab{b}})Lu, Qiu, Chang, Wu, Zhu, Rajpurohit, Clark, and Kalyan]{tabmwp}
Lu, P., Qiu, L., Chang, K.-W., Wu, Y.~N., Zhu, S.-C., Rajpurohit, T., Clark, P., and Kalyan, A.
\newblock Dynamic prompt learning via policy gradient for semi-structured mathematical reasoning.
\newblock \emph{arXiv preprint arXiv:2209.14610}, 2022{\natexlab{b}}.

\bibitem[Lu et~al.(2024{\natexlab{b}})Lu, Li, Chen, Xu, Luo, Zhang, and Ye]{ovis}
Lu, S., Li, Y., Chen, Q.-G., Xu, Z., Luo, W., Zhang, K., and Ye, H.-J.
\newblock Ovis: Structural embedding alignment for multimodal large language model.
\newblock \emph{arXiv preprint arXiv:2405.20797}, 2024{\natexlab{b}}.

\bibitem[Luan et~al.(2024)Luan, Feng, Chen, Wang, Zhou, and Li]{luan2024textcot}
Luan, B., Feng, H., Chen, H., Wang, Y., Zhou, W., and Li, H.
\newblock Textcot: Zoom in for enhanced multimodal text-rich image understanding.
\newblock \emph{arXiv preprint arXiv:2404.09797}, 2024.

\bibitem[Luo et~al.(2024)Luo, Liu, Liu, Phatale, Lara, Li, Shu, Zhu, Meng, Sun, et~al.]{omegaprm}
Luo, L., Liu, Y., Liu, R., Phatale, S., Lara, H., Li, Y., Shu, L., Zhu, Y., Meng, L., Sun, J., et~al.
\newblock Improve mathematical reasoning in language models by automated process supervision.
\newblock \emph{arXiv preprint arXiv:2406.06592}, 2024.

\bibitem[Masry et~al.(2022)Masry, Long, Tan, Joty, and Hoque]{masry2022chartqa}
Masry, A., Long, D.~X., Tan, J.~Q., Joty, S., and Hoque, E.
\newblock Chartqa: A benchmark for question answering about charts with visual and logical reasoning.
\newblock \emph{arXiv preprint arXiv:2203.10244}, 2022.

\bibitem[Mathew et~al.(2021)Mathew, Karatzas, and Jawahar]{mathew2021docvqa}
Mathew, M., Karatzas, D., and Jawahar, C.
\newblock Docvqa: A dataset for vqa on document images.
\newblock In \emph{Proceedings of the IEEE/CVF winter conference on applications of computer vision}, pp.\  2200--2209, 2021.

\bibitem[Mathew et~al.(2022)Mathew, Bagal, Tito, Karatzas, Valveny, and Jawahar]{infoqa}
Mathew, M., Bagal, V., Tito, R., Karatzas, D., Valveny, E., and Jawahar, C.
\newblock Infographicvqa.
\newblock In \emph{Proceedings of the IEEE/CVF Winter Conference on Applications of Computer Vision}, pp.\  1697--1706, 2022.

\bibitem[Methani et~al.(2020)Methani, Ganguly, Khapra, and Kumar]{methani2020plotqa}
Methani, N., Ganguly, P., Khapra, M.~M., and Kumar, P.
\newblock Plotqa: Reasoning over scientific plots.
\newblock In \emph{Proceedings of the IEEE/CVF Winter Conference on Applications of Computer Vision}, pp.\  1527--1536, 2020.

\bibitem[Mitra et~al.(2024)Mitra, Huang, Darrell, and Herzig]{CCot}
Mitra, C., Huang, B., Darrell, T., and Herzig, R.
\newblock Compositional chain-of-thought prompting for large multimodal models.
\newblock In \emph{Proceedings of the IEEE/CVF Conference on Computer Vision and Pattern Recognition}, pp.\  14420--14431, 2024.

\bibitem[OpenAI(2024)]{openai2024o1}
OpenAI.
\newblock Introducing openai o1, 2024.
\newblock URL \url{https://openai.com/o1/}.

\bibitem[Pitanov et~al.(2023)Pitanov, Skrynnik, Andreychuk, Yakovlev, and Panov]{pitanov2023monte}
Pitanov, Y., Skrynnik, A., Andreychuk, A., Yakovlev, K., and Panov, A.
\newblock Monte-carlo tree search for multi-agent pathfinding: Preliminary results.
\newblock In \emph{International Conference on Hybrid Artificial Intelligence Systems}, pp.\  649--660. Springer, 2023.

\bibitem[Qiao et~al.(2018)Qiao, Shen, Zhang, Wang, and Yuille]{qiao2018deep}
Qiao, S., Shen, W., Zhang, Z., Wang, B., and Yuille, A.
\newblock Deep co-training for semi-supervised image recognition.
\newblock In \emph{Proceedings of the european conference on computer vision (eccv)}, pp.\  135--152, 2018.

\bibitem[Saito et~al.(2018)Saito, Watanabe, Ushiku, and Harada]{saito2018maximum}
Saito, K., Watanabe, K., Ushiku, Y., and Harada, T.
\newblock Maximum classifier discrepancy for unsupervised domain adaptation.
\newblock In \emph{Proceedings of the IEEE Conference on Computer Vision and Pattern Recognition}, pp.\  3723--3732, 2018.

\bibitem[Schwenk et~al.(2022)Schwenk, Khandelwal, Clark, Marino, and Mottaghi]{A-okvqa}
Schwenk, D., Khandelwal, A., Clark, C., Marino, K., and Mottaghi, R.
\newblock A-okvqa: A benchmark for visual question answering using world knowledge.
\newblock In \emph{European conference on computer vision}, pp.\  146--162. Springer, 2022.

\bibitem[Seo et~al.(2015)Seo, Hajishirzi, Farhadi, Etzioni, and Malcolm]{geos}
Seo, M., Hajishirzi, H., Farhadi, A., Etzioni, O., and Malcolm, C.
\newblock Solving geometry problems: Combining text and diagram interpretation.
\newblock In \emph{Proceedings of the 2015 conference on empirical methods in natural language processing}, pp.\  1466--1476, 2015.

\bibitem[Shi et~al.(2024)Shi, Hu, Bin, Liu, Yang, Ng, Bing, and Lee]{math-llava}
Shi, W., Hu, Z., Bin, Y., Liu, J., Yang, Y., Ng, S.-K., Bing, L., and Lee, R. K.-W.
\newblock Math-llava: Bootstrapping mathematical reasoning for multimodal large language models.
\newblock \emph{arXiv preprint arXiv:2406.17294}, 2024.

\bibitem[Silver et~al.(2017)Silver, Schrittwieser, Simonyan, Antonoglou, Huang, Guez, Hubert, Baker, Lai, Bolton, et~al.]{silver2017mastering}
Silver, D., Schrittwieser, J., Simonyan, K., Antonoglou, I., Huang, A., Guez, A., Hubert, T., Baker, L., Lai, M., Bolton, A., et~al.
\newblock Mastering the game of go without human knowledge.
\newblock \emph{nature}, 550\penalty0 (7676):\penalty0 354--359, 2017.

\bibitem[Singh et~al.(2019)Singh, Natarajan, Shah, Jiang, Chen, Batra, Parikh, and Rohrbach]{textvqa}
Singh, A., Natarajan, V., Shah, M., Jiang, Y., Chen, X., Batra, D., Parikh, D., and Rohrbach, M.
\newblock Towards vqa models that can read.
\newblock In \emph{Proceedings of the IEEE/CVF conference on computer vision and pattern recognition}, pp.\  8317--8326, 2019.

\bibitem[Sun \& Jin(2011)Sun and Jin]{sun2011robust}
Sun, S. and Jin, F.
\newblock Robust co-training.
\newblock \emph{International Journal of Pattern Recognition and Artificial Intelligence}, 25\penalty0 (07):\penalty0 1113--1126, 2011.

\bibitem[Tong et~al.(2024)Tong, Brown, Wu, Woo, Middepogu, Akula, Yang, Yang, Iyer, Pan, et~al.]{Cambrian-1}
Tong, S., Brown, E., Wu, P., Woo, S., Middepogu, M., Akula, S.~C., Yang, J., Yang, S., Iyer, A., Pan, X., et~al.
\newblock Cambrian-1: A fully open, vision-centric exploration of multimodal llms.
\newblock \emph{arXiv preprint arXiv:2406.16860}, 2024.

\bibitem[Vagadia et~al.(2024)Vagadia, Chopra, Barnawal, Banerjee, Tuli, Chakraborty, and Paul]{vagadia2024phyplan}
Vagadia, H., Chopra, M., Barnawal, A., Banerjee, T., Tuli, S., Chakraborty, S., and Paul, R.
\newblock Phyplan: Compositional and adaptive physical task reasoning with physics-informed skill networks for robot manipulators.
\newblock \emph{arXiv preprint arXiv:2402.15767}, 2024.

\bibitem[Wang et~al.(2024{\natexlab{a}})Wang, Pan, Shi, Lu, Zhan, and Li]{mathvision}
Wang, K., Pan, J., Shi, W., Lu, Z., Zhan, M., and Li, H.
\newblock Measuring multimodal mathematical reasoning with math-vision dataset.
\newblock \emph{arXiv preprint arXiv:2402.14804}, 2024{\natexlab{a}}.

\bibitem[Wang et~al.(2024{\natexlab{b}})Wang, Bai, Tan, Wang, Fan, Bai, Chen, Liu, Wang, Ge, et~al.]{qwen2vl}
Wang, P., Bai, S., Tan, S., Wang, S., Fan, Z., Bai, J., Chen, K., Liu, X., Wang, J., Ge, W., et~al.
\newblock Qwen2-vl: Enhancing vision-language model's perception of the world at any resolution.
\newblock \emph{arXiv preprint arXiv:2409.12191}, 2024{\natexlab{b}}.

\bibitem[Wei et~al.(2022)Wei, Wang, Schuurmans, Bosma, Xia, Chi, Le, Zhou, et~al.]{wei2022chain}
Wei, J., Wang, X., Schuurmans, D., Bosma, M., Xia, F., Chi, E., Le, Q.~V., Zhou, D., et~al.
\newblock Chain-of-thought prompting elicits reasoning in large language models.
\newblock \emph{Advances in neural information processing systems}, 35:\penalty0 24824--24837, 2022.

\bibitem[Wu et~al.(2024)Wu, Chen, Pan, Liu, Liu, Dai, Gao, Ma, Wu, Wang, et~al.]{deepseek-vl2}
Wu, Z., Chen, X., Pan, Z., Liu, X., Liu, W., Dai, D., Gao, H., Ma, Y., Wu, C., Wang, B., et~al.
\newblock Deepseek-vl2: Mixture-of-experts vision-language models for advanced multimodal understanding.
\newblock \emph{arXiv preprint arXiv:2412.10302}, 2024.

\bibitem[Xie et~al.(2024)Xie, Goyal, Zheng, Kan, Lillicrap, Kawaguchi, and Shieh]{xie2024monte}
Xie, Y., Goyal, A., Zheng, W., Kan, M.-Y., Lillicrap, T.~P., Kawaguchi, K., and Shieh, M.
\newblock Monte carlo tree search boosts reasoning via iterative preference learning.
\newblock \emph{arXiv preprint arXiv:2405.00451}, 2024.

\bibitem[Xu et~al.(2024)Xu, Jin, Hao, Song, Sun, and Yuan]{llava-cot}
Xu, G., Jin, P., Hao, L., Song, Y., Sun, L., and Yuan, L.
\newblock Llava-o1: Let vision language models reason step-by-step.
\newblock \emph{arXiv preprint arXiv:2411.10440}, 2024.

\bibitem[Yang(2023)]{yang2023integrated}
Yang, F.
\newblock An integrated framework integrating monte carlo tree search and supervised learning for train timetabling problem.
\newblock \emph{arXiv preprint arXiv:2311.00971}, 2023.

\bibitem[Yao et~al.(2024{\natexlab{a}})Yao, Wu, Yang, Song, Zhang, Feng, Sun, Li, Ouyang, and Wang]{denseconnector}
Yao, H., Wu, W., Yang, T., Song, Y., Zhang, M., Feng, H., Sun, Y., Li, Z., Ouyang, W., and Wang, J.
\newblock Dense connector for mllms.
\newblock \emph{arXiv preprint arXiv:2405.13800}, 2024{\natexlab{a}}.

\bibitem[Yao et~al.(2024{\natexlab{b}})Yao, Yu, Zhao, Shafran, Griffiths, Cao, and Narasimhan]{yao2024tree}
Yao, S., Yu, D., Zhao, J., Shafran, I., Griffiths, T., Cao, Y., and Narasimhan, K.
\newblock Tree of thoughts: Deliberate problem solving with large language models.
\newblock \emph{Advances in Neural Information Processing Systems}, 36, 2024{\natexlab{b}}.

\bibitem[Yao et~al.(2024{\natexlab{c}})Yao, Yu, Zhang, Wang, Cui, Zhu, Cai, Li, Zhao, He, et~al.]{minicpm-v}
Yao, Y., Yu, T., Zhang, A., Wang, C., Cui, J., Zhu, H., Cai, T., Li, H., Zhao, W., He, Z., et~al.
\newblock Minicpm-v: A gpt-4v level mllm on your phone.
\newblock \emph{arXiv preprint arXiv:2408.01800}, 2024{\natexlab{c}}.

\bibitem[Ye et~al.(2021)Ye, Liu, Kurutach, Abbeel, and Gao]{ye2021mastering}
Ye, W., Liu, S., Kurutach, T., Abbeel, P., and Gao, Y.
\newblock Mastering atari games with limited data.
\newblock \emph{Advances in neural information processing systems}, 34:\penalty0 25476--25488, 2021.

\bibitem[Yu et~al.(2011)Yu, Krishnapuram, Rosales, and Rao]{yu2011bayesian}
Yu, S., Krishnapuram, B., Rosales, R., and Rao, R.~B.
\newblock Bayesian co-training.
\newblock \emph{The Journal of Machine Learning Research}, 12:\penalty0 2649--2680, 2011.

\bibitem[Yu et~al.(2024)Yu, Yao, Zhang, He, Han, Cui, Hu, Liu, Zheng, Sun, et~al.]{rlhf-V}
Yu, T., Yao, Y., Zhang, H., He, T., Han, Y., Cui, G., Hu, J., Liu, Z., Zheng, H.-T., Sun, M., et~al.
\newblock Rlhf-v: Towards trustworthy mllms via behavior alignment from fine-grained correctional human feedback.
\newblock In \emph{Proceedings of the IEEE/CVF Conference on Computer Vision and Pattern Recognition}, pp.\  13807--13816, 2024.

\bibitem[Yue et~al.(2024)Yue, Zheng, Ni, Wang, Zhang, Tong, Sun, Yu, Zhang, Sun, et~al.]{mmmu_pro}
Yue, X., Zheng, T., Ni, Y., Wang, Y., Zhang, K., Tong, S., Sun, Y., Yu, B., Zhang, G., Sun, H., et~al.
\newblock Mmmu-pro: A more robust multi-discipline multimodal understanding benchmark.
\newblock \emph{arXiv preprint arXiv:2409.02813}, 2024.

\bibitem[Zelikman et~al.(2022)Zelikman, Wu, Mu, and Goodman]{zelikman2022star}
Zelikman, E., Wu, Y., Mu, J., and Goodman, N.
\newblock Star: Bootstrapping reasoning with reasoning.
\newblock \emph{Advances in Neural Information Processing Systems}, 35:\penalty0 15476--15488, 2022.

\bibitem[Zhang et~al.(2024{\natexlab{a}})Zhang, Zhoubian, Hu, Yue, Dong, and Tang]{zhang2024rest}
Zhang, D., Zhoubian, S., Hu, Z., Yue, Y., Dong, Y., and Tang, J.
\newblock Rest-mcts*: Llm self-training via process reward guided tree search.
\newblock \emph{arXiv preprint arXiv:2406.03816}, 2024{\natexlab{a}}.

\bibitem[Zhang et~al.(2024{\natexlab{b}})Zhang, Gao, Gan, Dufter, Wenzel, Huang, Shah, Du, Zhang, Li, et~al.]{mm1.5}
Zhang, H., Gao, M., Gan, Z., Dufter, P., Wenzel, N., Huang, F., Shah, D., Du, X., Zhang, B., Li, Y., et~al.
\newblock Mm1. 5: Methods, analysis \& insights from multimodal llm fine-tuning.
\newblock \emph{arXiv preprint arXiv:2409.20566}, 2024{\natexlab{b}}.

\bibitem[Zhang et~al.(2024{\natexlab{c}})Zhang, Huang, Jin, and Lu]{vision_survey}
Zhang, J., Huang, J., Jin, S., and Lu, S.
\newblock Vision-language models for vision tasks: A survey.
\newblock \emph{IEEE Transactions on Pattern Analysis and Machine Intelligence}, 2024{\natexlab{c}}.

\bibitem[Zhang et~al.(2024{\natexlab{d}})Zhang, Zhang, Li, Zhang, Sun, Gan, Yang, Pang, and Yang]{llava-reasoner}
Zhang, R., Zhang, B., Li, Y., Zhang, H., Sun, Z., Gan, Z., Yang, Y., Pang, R., and Yang, Y.
\newblock Improve vision language model chain-of-thought reasoning.
\newblock \emph{arXiv preprint arXiv:2410.16198}, 2024{\natexlab{d}}.

\bibitem[Zhang et~al.(2023)Zhang, Wu, Zhao, Lin, Zhang, Wang, and Xie]{pmcvqa}
Zhang, X., Wu, C., Zhao, Z., Lin, W., Zhang, Y., Wang, Y., and Xie, W.
\newblock Pmc-vqa: Visual instruction tuning for medical visual question answering.
\newblock \emph{arXiv preprint arXiv:2305.10415}, 2023.

\bibitem[Zhao et~al.(2022)Zhao, Li, Li, and Zhang]{zhao2022multihiertt}
Zhao, Y., Li, Y., Li, C., and Zhang, R.
\newblock Multihiertt: Numerical reasoning over multi hierarchical tabular and textual data.
\newblock \emph{arXiv preprint arXiv:2206.01347}, 2022.

\end{thebibliography}
\bibliographystyle{icml2025}

\newpage
\clearpage
\appendix
\onecolumn
\section{The Sources of Raw Data}
\label{app: data sources}

To construct a comprehensive and general-purpose tree-based reasoning dataset, we collect 260K raw multimodal input questions spanning varouis domain, including 
\begin{itemize}
    \item 55K Mathematical Data: From GLLaVA~\cite{g-llava}, GEOS~\cite{geos}, UniGeo~\cite{chen2022unigeo}, GeoQA Plus~\cite{chen-etal-2021-geoqa}, Geo3K~\cite{geo3k}, MathVision~\cite{mathvision}, GeoMverse~\cite{geomverse}, and MathV360K~\cite{math-llava}.
    \item 116K Figure Understanding data: From DVQA~\cite{kafle2018dvqa}, DocVQA~\cite{mathew2021docvqa}, FigureQA~\cite{kahou2017figureqa}, PlotQA~\cite{methani2020plotqa}, ChartQA~\cite{masry2022chartqa}, InfoVQA~\cite{infoqa}, MultiHiertt~\cite{zhao2022multihiertt}, and LRV-Chart~\cite{lrv_chart}.
    \item 41K Math Word Problem Data: From IconQA~\cite{lu2021iconqa}, TabMWP~\cite{tabmwp}, CLEVR~\cite{johnson2017clevr}, CLEVR-Math~\cite{clevr-math}, and Super-CLEVR~\cite{super-clevr}.
    \item 2K Mdeical Data: From VQA-RAD~\cite{VQA-RAD}, and PMC-VQA~\cite{pmcvqa}.
    \item 17K Sience Data: From TQA~\cite{tqa}, AI2D~\cite{ai2d}, and ScienceQA~\cite{scienceqa}.
    \item 24K Nature World QA Data: From VQA-AS~\cite{vqa-as}, A-OKVQA~\cite{A-okvqa}, TextVQA~\cite{textvqa}, Vizwiz~\cite{gurari2018vizwiz}, and VQA2.0~\cite{vqa2.0}.
\end{itemize}

\end{document}